\documentclass[journal]{IEEEtran}
\usepackage{cite}
\usepackage{amsmath,amssymb,amsfonts,amsthm,upgreek}
\usepackage{tabularx}
\usepackage{comment}
\usepackage{graphicx}
\usepackage{float}
\usepackage{subcaption}
\usepackage{caption}  
\usepackage{multirow} 
\usepackage{textcomp}
\usepackage{afterpage}
\usepackage{lipsum}
\usepackage{array}
\usepackage{booktabs}
\usepackage{adjustbox}
\usepackage[colorlinks=true, linkcolor=blue, citecolor=blue, urlcolor=blue]{hyperref}
\usepackage[utf8]{inputenc}
\usepackage[linesnumbered,ruled,vlined]{algorithm2e} 
\usepackage{colortbl}
\usepackage{soul,color}
\usepackage{enumitem}
\usepackage{xcolor}
\usepackage{multicol}
\usepackage{threeparttable}

\DeclareMathOperator*{\argmax}{arg\,max}

\ifCLASSINFOpdf
\else
\fi
\begin{document}
%

\title{Dynamic Graph-Like Learning with Contrastive Clustering on Temporally-Factored Ship Motion Data for Imbalanced Sea State Estimation in Autonomous Vessel}
%
%
%

\author{Kexin Wang,
        Mengna Liu,
        Xu Cheng,~\IEEEmembership{Member,~IEEE,}
        Fan Shi,
        Shanshan Qi,
        and~Shengyong Chen,~\IEEEmembership{Senior Member,~IEEE,}
\thanks{All authors are with the School of Computer Science and Engineering, Tianjin University of Technology, Tianjin, China. This work has been submitted as a preprint to arXiv.}
}

\maketitle

\begin{abstract}

Accurate sea state estimation is crucial for the real-time control and future state prediction of autonomous vessels. However, traditional methods struggle with challenges such as data imbalance and feature redundancy in ship motion data, limiting their effectiveness. To address these challenges, we propose the Temporal-Graph Contrastive Clustering Sea State Estimator (TGC-SSE), a novel deep learning model that combines three key components: a time dimension factorization module to reduce data redundancy, a dynamic graph-like learning module to capture complex variable interactions, and a contrastive clustering loss function to effectively manage class imbalance. Our experiments demonstrate that TGC-SSE significantly outperforms existing methods across 14 public datasets, achieving the highest accuracy in 9 datasets, with a 20.79\% improvement over EDI. Furthermore, in the field of sea state estimation, TGC-SSE surpasses five benchmark methods and seven deep learning models. Ablation studies confirm the effectiveness of each module, demonstrating their respective roles in enhancing overall model performance. Overall, TGC-SSE not only improves the accuracy of sea state estimation but also exhibits strong generalization capabilities, providing reliable support for autonomous vessel operations.

\end{abstract}


%
\IEEEpeerreviewmaketitle

\section{Introduction}
%
%
%
%

\IEEEPARstart{E}{fficient} and secure maritime operations are vital for mitigating climate change, ensuring industry sustainability, and maintaining cost-efficiency \cite{wang2023green}. The significance of these operations is especially pronounced in autonomous ships, which are considered central to future low-carbon solutions \cite{zis2023design}. These vessels uniquely confront the challenge of navigating unpredictable sea conditions \cite{cheng2023novel}. To address this issue, increasing attention is being paid to sea state estimation (SSE), which involves the comprehensive analysis and classification of marine characteristics such as wave height, period, and direction. This approach aims to improve autonomous ship control systems and facilitate decision-making under hazardous conditions \cite{liu2024prototype}.

SSE traditionally relies on either subjective human observations or advanced instruments, but human errors are the primary cause of accidents \cite{maternova2023human}. Additionally, each type of instrument has inherent limitations: wave buoys have limited coverage, and although wave radars provide continuous accuracy under various weather conditions, they require complex processing and precise calibration \cite{stredulinsky2011ship}. In response to these challenges, researchers have proposed the Wave Buoy Analog (WBA) method. This approach treats ships as mobile wave buoys, utilizing onboard sensors to analyze ship movements and estimate sea states through model-based or data-driven techniques.

The model-based approach to understanding ocean wave dynamics employs sophisticated physical and mathematical models to elucidate the intricate processes affecting wave behavior and predict changes in sea conditions. This method adeptly addresses the non-linear and stochastic nature of waves, incorporating their interactions with environmental elements such as wind, currents, and tides. While the model-based approach is effective in capturing the non-linear and stochastic nature of ocean waves, it also presents significant challenges. These challenges include the need for specialized knowledge to develop and solve the models, as well as the complexity of accurately determining model parameters and boundary conditions.

In the field of SSE, data-driven methods leverage historical and real-time ship motion data to discern the correlation between sea states and ship dynamics, which aids in the classification of sea states. This approach relies on large amounts of high-quality data rather than complex physical or mathematical models, and does not require specialized prior knowledge. Existing implementations of this method include Machine Learning (ML) models employing multi-domain handcrafted features \cite{9561261}, hybrid models integrating Long Short-Term Memory (LSTM), Convolutional Neural Networks (CNN), and Fast Fourier Transform (FFT) processes \cite{8794069}, as well as densely connected CNN \cite{8794069}.

Despite the notable advancements in data-driven models for SSE, several limitations persist:

\begin{enumerate}
 \item Ship motion data is typical time series data. A profound challenge in the domain of time series is how to overcome information redundancy and observe the underlying patterns, thereby improving the model's ability to discern subtle temporal dynamics. 

  \item The underappreciated correlations within a ship's six degrees of freedom data present a critical challenge for SSE modeling. Previous models have often overlooked the complex interplay between these variables, leading to incomplete and potentially misleading representations of a vessel's dynamic behavior. Addressing this challenge requires innovative approaches that can capture and model these intricate relationships to support more accurate and nuanced navigational predictions.

 \item A significant hurdle in maritime data analysis is the pronounced class imbalance, as depicted in Fig. \ref{fig:1}. The prevalence of certain sea states over others skews the model's learning process, particularly in regions where data scarcity is exacerbated by extreme weather events. This imbalance not only hampers the model's generalization capabilities but also poses a threat to its reliability in predicting rare yet critical sea states, which are essential for robust maritime operations.
 
 \begin{figure} [htp]
        \centering
		\includegraphics[width=0.44\textwidth]{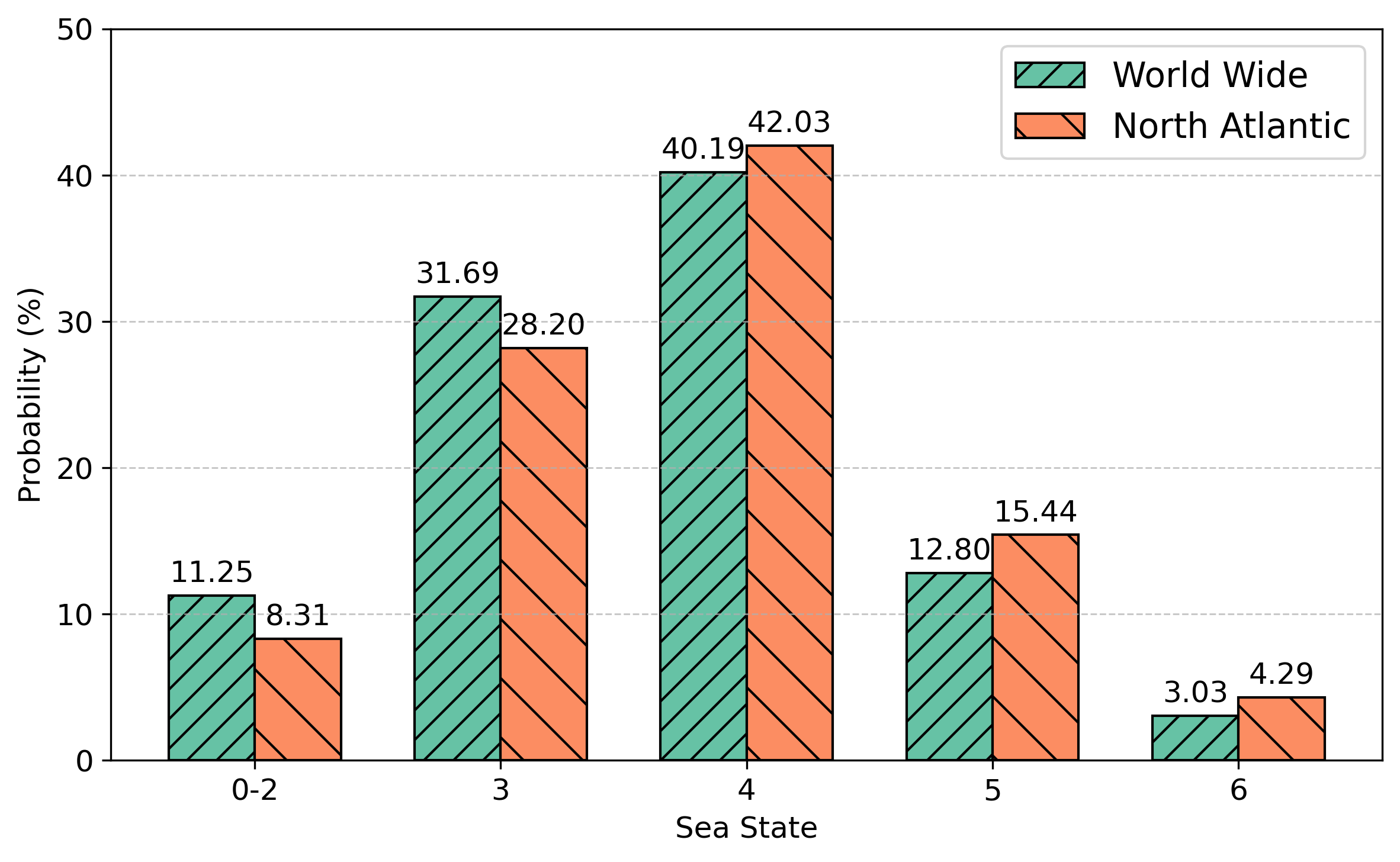}
		\vspace{-5pt}
        	\caption{Imbalanced Sea State Probability Distribution for Worldwide and North Atlantic.}
		\label{fig:1}
		\vspace{-5pt}
\end{figure}

\end{enumerate}

To address the above challenges, we propose a novel data-driven model called Temporal-Graph Contrastive Sea State Estimator (TGC-SSE). The TGC-SSE model integrates three key components, each of which is customised to address a specific challenge: First, we design a time dimension decomposition module that specifically targets the problem of information redundancy in time series data and reveals the underlying patterns. By performing slicing, sampling, feature extraction, and merging operations on the time dimension, the module effectively captures temporal dependencies, reduces data redundancy, and improves the computational efficiency of the model. Next, to overcome the under-appreciated correlation challenge in vessel six-degree-of-freedom data, we introduce a dynamic graph-like learning module. This module extracts correlation information between different variables at the same point in time and constructs a dynamic graph structure based on the interaction strength of the variables' features, thus capturing the interrelationships between the variables and supporting more accurate voyage prediction. Finally, to address the significant class imbalance in maritime data analysis, we introduce a contrast clustering loss function. This function integrates intra-class compactness, inter-class divisibility, and category distribution balance, which effectively alleviates the category imbalance problem and improves the generalisation ability of the model. The specific contributions of this paper are as follows:

\begin{enumerate}
\item  We present a dual-module approach within the TGC-SSE that adeptly addresses the complexities of ship motion data. The Time Dimension Factorization Module captures temporal dynamics and channel dependencies, enhancing computational efficiency through strategic downsampled sequence generation and slicing operations. Concurrently, the Dynamic Graph-like Learning Module delves into the intricate inter-variable relationships, enriching the model's predictive accuracy with a nuanced representation of a vessel's behavior.

\item The proposed class imbalance loss function enhances model performance by emphasizing intraclass compactness, interclass separability, and class distribution balance. By combining positive and negative pairwise comparison losses with a clustering distribution loss, this dual method promotes balanced clustering and better class separability, effectively mitigating class imbalance issues in SSE.

\item The TGC-SSE model is subjected to rigorous evaluation against state-of-the-art (SOTA) time series classification methods on public UEA (University of East Anglia) datasets. Furthermore, it is benchmarked against leading SSE approaches and class imbalance learning techniques on two bespoke ship motion datasets. The significance of the proposed modules and pivotal parameters is validated through meticulous ablation and sensitivity analyses, underscoring the robustness and adaptability of our model.

\end{enumerate}

This paper is structured as follows: Section \ref{Relatedwork} reviews pertinent literature. Section \ref{method} outlines the problem definition and details our proposed model. In Section \ref{experiments}, we showcase our model's performance and advantages through experimental results. Finally, Section \ref{conclusion} summarizes key findings, implications, and potential future research directions.

\begin{figure*}[t!]
\centering
\includegraphics[width=1.0\textwidth]{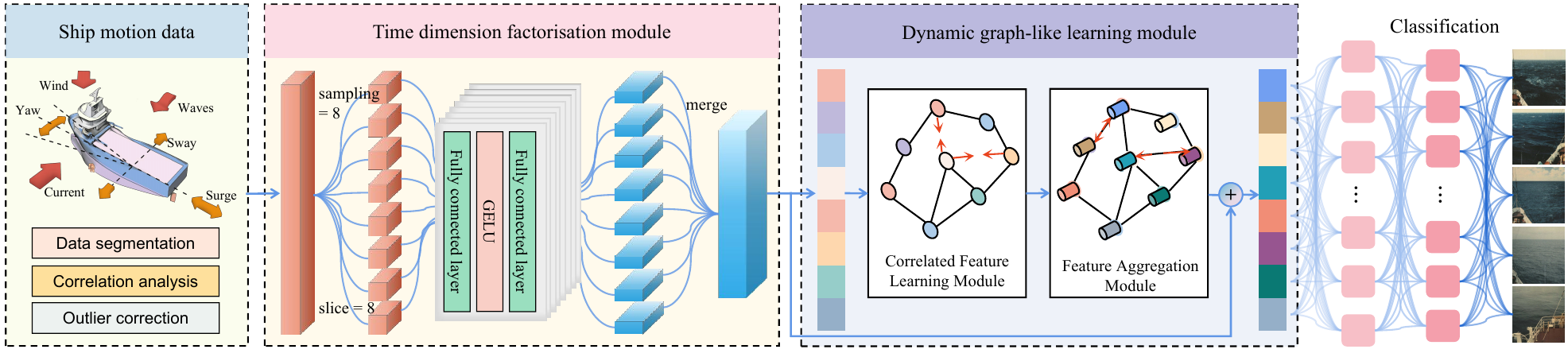}

\caption{The pipeline of the proposed model.}
\label{fig:structure}
\end{figure*}

\section{Related Work}
\label{Relatedwork}
In this comprehensive review, we critically analyze several SOTA techniques used for estimating sea state, encompassing both model-based and data-driven approaches.

\subsection{Model-based Approach}
Model-based approaches involve the parametrization or Bayesian modeling of wave spectra based on certain assumptions or prior knowledge. One example is the brute-force spectral method introduced by Nielsen et al. \cite{nielsen2018brute}, which relies on the wave buoy analogy. This method estimates the wave spectrum using a direct and straightforward calculation, making it efficient. However, this approach requires validation across various ship types and sensitivity analyses. Ren et al. propose alternative techniques suitable for different ships and SSE requirements, utilizing Bessel surfaces and L1 optimization to provide smooth, robust, and scalable solutions \cite{REN2021102904}. Nonetheless, these methods are not applicable to ships with forward speed. Nielsen's 15 years of research has focused on wave buoy analogies, while Kalman filtering, a well-known and efficient prediction algorithm, is widely used across different fields \cite{NIELSEN2017352}. Similarly, Loizou et al. integrate Kalman filtering with least squares approximation to estimate uncalibrated video amplitude spectra, converting them to meters for wave height estimation. The applicability of this method to higher sea states and its reliance on local ocean video amplitude spectrum calibration remain uncertain \cite{loizou2021sea}. In summary, precise modeling of the complex and dynamic ocean environment is essential for the success of model-based approaches.

\subsection{Data-driven Approach}
The data-driven approach utilizes shipboard sensors to extract temporal and frequency features from continuous time-series data. Techniques like classification, regression, and clustering identify patterns and explore relationships in categorical information \cite{MAJIDIAN2022111684}. ML algorithms have gained popularity in SSE due to their ability without the need for an accurate motion dynamic model. Han et al. proposed an initial approach that addressed the challenge of linking wave and ship motions using statistical and wavelet analysis. By constructing multi-domain features and integrated ML models, they overcame this challenge, but only for data collected near the shore \cite{9561261}. Subsequently, they introduced a hybrid approach that combines the WBA method with ML to compensate for lacking sea state information in training data. However, the accuracy of uncertainty representation and reliance on the model-based approach are critical factors \cite{9405466}.

In addition to the methods discussed earlier, recent research has explored the application of deep learning (DL) techniques to sea state estimation (SSE). DL, a subset of machine learning (ML), utilizes neural networks for SSE, distinguishing it from traditional ML approaches. Notably, Cheng et al. proposed SeaStateNet in 2019, which combined LSTM, CNN, and FFT to achieve promising results \cite{8794069}. Building on this work, Cheng et al. later developed SSENET, which leverages hierarchical features from raw time-series sensor data (e.g., wave height and direction) to estimate sea states \cite{8962178}. Also in 2019, SpectralNet was introduced, using short-time Fourier transform (STFT) to convert ship data into spectrograms. These spectrograms are then fed into a two-dimensional CNN for sea state classification, resulting in improved accuracy compared to methods using raw data \cite{9254890}. In 2023, Cheng et al. identified a new challenge in SSE: class imbalance in ship motion datasets across different sea areas, and proposed using a prototypical classifier to address this issue \cite{cheng2023novel}. Li et al. introduced SecureSSE \cite{10273201}, a robust adversarial method designed to mitigate the impact of data attacks in sea state estimation. Despite the significant advancements of these methods, existing approaches have not fully addressed the challenges of separately extracting features from the time and channel dimensions to reduce feature redundancy, nor have they explored whether the features of ship motion data are correlated.

\section{Temporal-Graph Contrastive Clustering SSE}
\label{method}
\subsection{Problem Definition}
\label{AA}

In this study, we approach SSE as a multivariate classification task with temporal characteristics. Our main objective is to classify different sea states using data collected by Inertial Measurement Units (IMUs) sensors placed on board. We classify the sea state into ten categories based on wave heights, as shown in the table \ref{table_seacode}. Notably, our analysis focuses on the first seven sea states, which account for 98\% of all recorded sea states, ranging from calm (state 0) to moderate conditions (state 6). We grouped sea states 0 to 2 together because they all indicate relatively calm seas. However, we excluded sea states 7 to 9 from our analyses because they are rare and usually indicate conditions that are unsafe for shipping. It is worth noting that we used ship motion data as inputs to the model rather than wave heights, among other reasons, because waves are categorised as regular or irregular and it is difficult to monitor wave heights in real time when irregular waves are encountered.

\begin{table}[htp]
\captionsetup{font=small}
\caption{Definition of Sea State \cite{fossen2011handbook,WAN2022110600}}
\centering
\label{table_seacode}
\resizebox{0.96\linewidth}{!}{%
\begin{tabular}{cccccc}
\toprule
\multirow{2}{*}{\textbf{Sea State}} & \multirow{2}{*}{\textbf{Description}} & \multirow{2}{*}{\textbf{Wave Height}} & \multicolumn{2}{c}{\textbf{Probability}} \\
\cmidrule{4-5}
& & & \textbf{Worldwide} & \textbf{North Atlantic} \\
\midrule
0 & Calm (glassy) & 0 & --- & --- \\
1 & Calm (ripples) & 0-0.1 & 11.2486 & 8.3103 \\
2 & Smooth & 0.1-0.5 & --- & --- \\
3 & Slight & 0.5-1.25 & 31.6851 & 28.1996 \\
4 & Moderate & 1.25-2.5 & 40.1944 & 42.0273 \\
5 & Rough & 2.5-4.0 & 12.8005 & 15.4435 \\
6 & Very rough & 4.0-6.0 & 3.0253 & 4.2938 \\
7 & High & 6.0-9.0 & 0.9263  & 1.4968 \\
8 & Very High & 9.0-14.0 & 0.1190 & 0.2263 \\
9 & Extreme & $>$ 14.0 & 0.0009  & 0.0016 \\
\bottomrule
\end{tabular}%
}
\end{table}

We denote ship motion data as \( \mathcal{X} = [X_1, X_2, \ldots, X_N] \in \mathbb{R}^{N \times F \times T} \), where \( N \) represents the number of time windows. Each time window \( X_{i} = [X_{i,1}, X_{i,2}, \ldots, X_{i,T}] \in \mathbb{R}^{F \times T} \) corresponds to the \( i \)-th time window, with \( F \) indicating the number of input features and \( T \) being the time series length. The sea state class denoted as \( y \in \{1, \ldots, 5\} \). Our objective is to develop a mapping function \( f : \mathcal{X} \rightarrow y \) that can precisely classify sea states from the ship motion data. To achieve this, we introduce an innovative deep learning (DL) model, the details of which are presented in subsequent subsections.

\subsection{Model Overview}
\label{TGC-SSE model}

The proposed SSE model, shown in Fig. \ref{fig:structure} and Fig. \ref{fig:probability}, consists of four key modules: the data preprocessing module, the time-dimension factorization module, the dynamical graph learning module, and the contrastive clustering loss function. The purpose of the data preprocessing module is to enhance the raw ship motion data by correcting anomalies, conducting Pearson correlation analysis, and employing a sliding window method to minimize sensor noise. Subsequently, the temporal dimension factorization module processes the ship motion data into interleaved sub-sequences through slicing and downsampling operations, where each sub-sequence is learned separately for temporal information, and then stitched together in the original order. The dynamic graph-like learning module focuses on capturing the interactions between dimensions on the channel dimension, allowing the model to learn and construct graph structures to extract more useful information to improve classification accuracy. Finally, the contrastive clustering loss addresses the class imbalance in ship motion data by adjusting training weights to ensure the classifier correctly identifies sea states.

\subsection{Time Dimension Factorization Module}

To address the redundancy inherent in time series data and extract multi-scale features, we employ the concept of group convolution \cite{10379602, 9811376} to decompose the original time series into sub-sequences. By learning the features of these sub-sequences, we can more effectively capture multi-scale features while reducing computational complexity. 

The temporal dynamics of ship motion are encapsulated within a tensor \( X \in \mathbb{R}^{F \times T} \), where \( F \) denotes the number of features and \( T \) the sequence length. A critical examination of the temporal windows revealed inherent temporal redundancy, which is evident in the high similarity between the original sequences and their corresponding slices.

To mitigate the impact of temporal redundancy, we introduce a Time Dimension Factorization Module that adeptly processes the temporal dimension prior to the dynamic graph-like learning phase. This module is designed to distill and refine the temporal features, enhancing the subsequent feature representation.

Our approach segments the time series data into \( s \) subsequences, leveraging the observed redundancy to reduce the effective sequence length without loss of critical information. Each subsequence is independently processed by a Multilayer Perceptron (MLP) to capture localized temporal dynamics. The slicing operation is mathematically represented as:
\begin{equation} 
{x_1},...,{x_s} = S(norm(X)).
\label{eq:1}
\end{equation} 
where \( \text{norm}(X) \) refers to the normalised tensor \( X \) and \( \text{S} \) is the operation of dividing the tensor into \( s \) interleaved subsequences.

Each subsequence \( x_i \) is then transformed into a feature representation \( X_i^T \) via the temporal feature extractor MLP:
\begin{equation} 
X_i^T = T({x_i}),i \in [1,s].
\label{eq:2}
\end{equation}
where \( T \) symbolizes the temporal feature extraction function implemented by the MLP.

The feature representations of the subsequences are subsequently merged to restore the temporal continuity:
\begin{equation} 
{X^T} = M(X_1^T,...,X_s^T).
\label{eq:3}
\end{equation} 
where \( M \) denotes the merging function that orders the features extracted from each subsequence chronologically.

To complement the temporal feature extraction, a one-dimensional Convolutional Neural Network (1D CNN) extracts additional features from the preprocessed data. These features are concatenated with the temporal features \( X^T \) to form a comprehensive feature set \( X_e \) :
\begin{equation} 
{X_e} = Cat({X^T},{X_{cnn}}).
\label{eq:4}
\end{equation} 
where \( X_{\text{cnn}} \) represents the features extracted by the 1D CNN, and \( \text{Cat} \) denotes the concatenation operation.

Here, \( X_{\text{cnn}} \) represents the features extracted by the 1D CNN. Although these features are not explicitly shown in the model diagram, this step is included in the actual experiments. The benefits of incorporating CNN were confirmed in the ablation experiments. The term \( \text{Cat} \) denotes the concatenation operation.

This integrated feature extraction process ensures that our model captures the nuanced temporal dynamics within the ship motion data, providing a rich and informative feature set for the dynamic graph-like learning module.

The choice of \( s \) is substantiated through a sensitivity analysis detailed in the experimental section, which demonstrates the impact of \( s \) on model performance and justifies our selection of \( s = 8 \).

\subsection{Dynamic Graph-like Learning Module}

Graph neural networks (GNNs) are very useful tools to capture interdependencies and interactions between different sensor signals, which can provide a more holistic view of the monitored system. But, GNNs encounter several limitations when applied to time series modeling. Primarily, GNNs are designed to process data with intrinsic graph structures, such as social networks or molecular configurations, rather than the sequential nature of time series data. This fundamental discrepancy can result in suboptimal performance, as GNNs may struggle to effectively capture and represent temporal dependencies. Furthermore, incorporating temporal information into GNNs often necessitates intricate architectures and substantial computational resources, leading to increased model complexity and extended training durations. Another significant limitation is the potential for poor generalization across diverse types of time series data, particularly when dealing with highly dynamic or non-stationary temporal patterns.

Therefore, a dynamic graph-like learning module is proposed by borrowing the idea of graph learning using the edge and node to describe the correlation of different signals. This module constructs a graph structure to model the dynamic interconnections between channels. Each node in the graph represents a channel feature, and each edge represents the relationship between a pair of channels. The module operates in three key steps: association learning, interaction strength learning, and dynamic aggregation. These steps enable the model to iteratively refine the graph representation, capturing complex feature associations and dependencies. The following sections elaborate on these steps:

\paragraph{Associated learning}

We convert the output of the time dimension factorization module, ${X_e}$, into edge features by performing a node-to-edge conversion. For each pair of nodes $(i, j)$, the edge features $e_{i, j}$ are obtained using the node-to-edge feature conversion function $f_{\text{n}\to\text{e}}( )$, which extracts the association between node $i$ and node $j$. Mathematically, the node-to-edge transformation is defined as follows

\begin{equation}
e_{i,j} = {f_{n \to e}}(Cat[{X_{e,i}},{X_{e,j}}]).
\label{eq:5}
\end{equation}

\paragraph{Interaction strength learning}

To evaluate the interaction strength between feature pairs, we discretely learn the graph structure using Eq. \ref{eq:interaction}. In this process, we designate certain edges as 'non-edges,' meaning these edges do not contribute to the transmission of information. Information is only propagated along the connected edges. The interaction strength between the \(i\)-th and \(j\)-th features, \(In_{i,j}\), is computed using the transformation function \(g(e_{i,j})\), followed by the application of a sigmoid function \cite{elfwing2018sigmoid}:

\begin{equation}
\label{eq:interaction}
In_{i,j} = \textit{sigmoid}(g(e_{i,j})) .
\end{equation}

This equation captures the interaction strength by transforming the edge information and applying a sigmoid activation, allowing us to quantify the relationships between connected features.

\paragraph{Dynamic aggregation}

In the dynamic aggregation phase, the features obtained from the first two steps are dynamically aggregated back into the two nodes of each edge using the edge features $e_{i,j}$ and the interaction strength \(In_{i,j}\). The following equation illustrates the process of aggregating the features of the edge connecting node $i$ and node $j$ back to node $i$:

\begin{equation}
\label{eq:7}
x_i^{(2)} = \sum_{j \neq i} e_{i,j} \cdot In_{i,j}.
\end{equation}
where $\mathbf{x}_i^{(2)}$ represents the aggregated feature of node $i$ after dynamic aggregation.

By repeating these steps across multiple iterations, the dynamic graph-like structure construction module progressively refines the graph representation, allowing the model to capture complex feature associations and dependencies. Although this module may incur higher computational costs compared to some baseline methods, we mitigate this by first using a 1D CNN to reduce the dimensionality of the feature extraction module's output, thereby decreasing the number of features that need to be processed by the dynamic graph-like structure construction module. Additionally, we use multiple iterations to progressively optimize the graph representation, instead of constructing the complete graph structure in one go, reducing both computational effort and memory consumption.

\subsection{Contrastive Clustering for Imbalanced Classification}
In traditional methods, a softmax classifier is utilized to determine the probability of each class. Considering $g(x) \in \mathbb{R}^{N \times F}$ and $V \in \mathbb{R}^{F \times k}$, where $g(x)$ denotes the features obtained from the previous two modules and $V$ represents the softmax classifier parameters, the classification procedure can be expressed as:

\begin{equation}
\ell = V^T g(x) = [v_1^T g(x); v_2^T g(x); \cdots; v_k^T g(x)],
\end{equation} 
where $N$ is the number of samples, $F$ is the number of input features, $k$ is the number of classes, and $v_i \in \mathbb{R}^F$ is the weight vector corresponding to the $i$-th class, with $i = 1, 2, \ldots, k$.

Next, the learning process is optimized using a cross-entropy loss function, $\mathcal{L}$, which aims to minimize the discrepancy between the predicted probabilities and the true class labels. It is defined as:
\begin{equation}
\label{loss}
\mathcal{L} = \sum_i^k \frac{N_i}{N} \mathcal{L}(D_i),
\end{equation}
here, \( N_i \) denotes the quantity of training samples for the \( i \)-th class, \( N \) represents the overall number of samples, and \( D_i \) signifies the set of samples associated with the \( i \)-th class.

Based on Eq. (\ref{loss}), in the presence of class imbalance where $N_1 \leq N_2 \leq \cdots \leq N_k$, it follows that $\mathcal{L}_1 \leq \mathcal{L}_2 \leq \cdots \leq \mathcal{L}_k$. Consequently, the decision boundary tends to favor the minority class.

\begin{figure} [htp]
        \centering
		\includegraphics[width=0.44\textwidth]{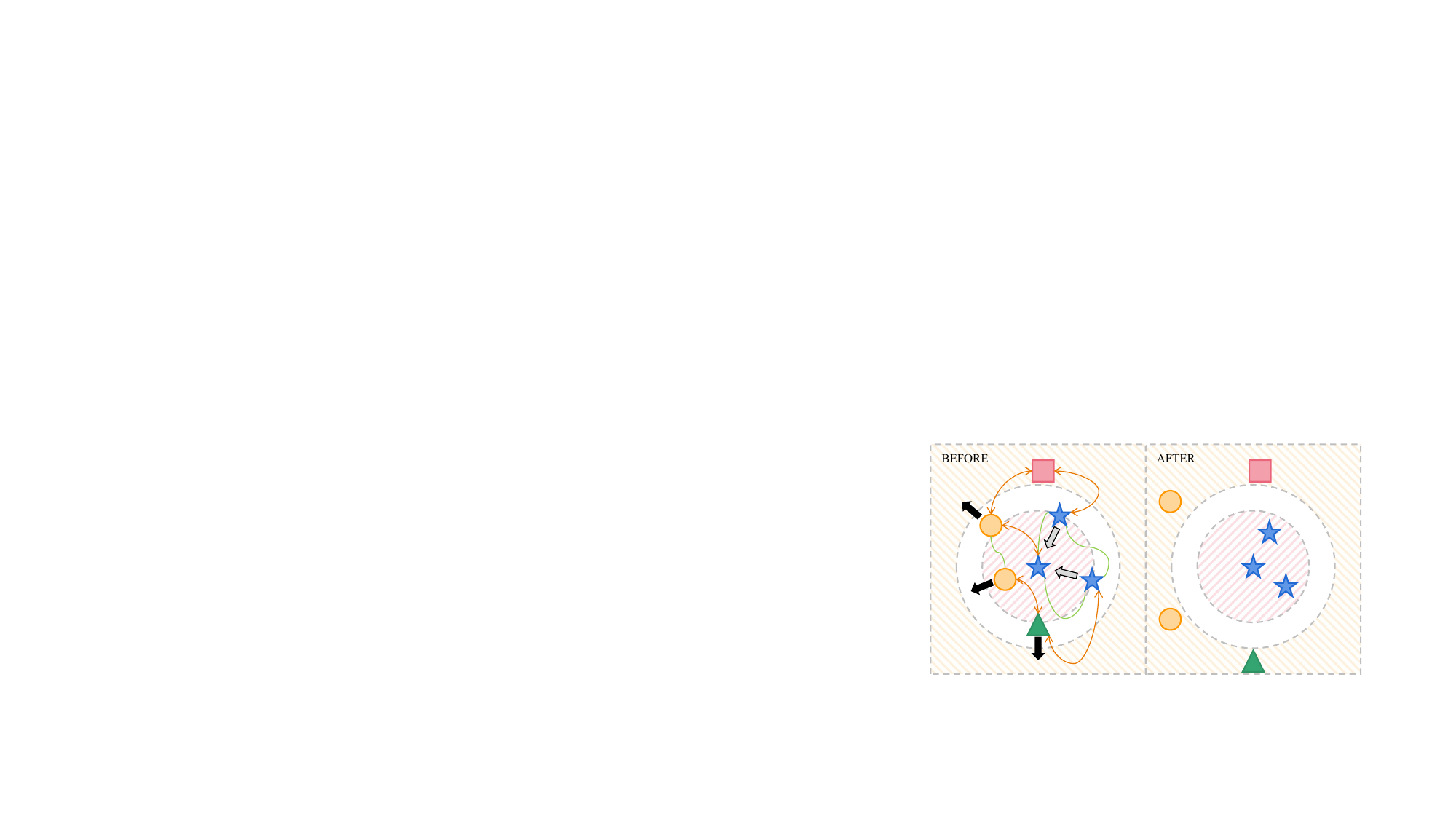}
		\vspace{-5pt}
        \caption{In the figure, the Contrastive clustering loss function is depicted. Each shape corresponds to a unique sea state class. During training, classes with a smaller sample size receive increased weighting. The thick arrow is denoted as ‘\( L_{\text{cluster}} \)’, the orange bi-directional arrow as ‘\( L_{\text{neg}} \)’, and the green line represents ‘\( L_{\text{pos}} \)’.}
		\label{fig:probability}
		\vspace{-5pt}
\end{figure}

In this study, we propose an alternative approach known as \textit{ContrastiveClusteringLoss} to tackle the class imbalance issue in DL frameworks. This loss function comprises four elements: feature normalization, contrastive loss for both positive and negative pairs, cluster assignment loss, and class weights to handle imbalanced data distribution. The schematic is shown in Fig. \ref{fig:probability}.

\paragraph{Feature Normalization}
The normalization process, represented by \( \hat{x}_i = \frac{x_i}{||x_i||_2} \), ensures that feature vectors are evaluated on a consistent scale, focusing on the directionality rather than the magnitude of the vectors. This step is foundational for the subsequent contrastive loss calculations.
\paragraph{Class Weights}
To address the class imbalance problem, we introduce class weights into the contrastive loss calculation. Class weights are assigned based on the proportion of each class in the dataset, ensuring that minority classes have a greater impact on the loss. This adjustment helps the model learn more effectively from underrepresented classes, mitigating bias towards majority classes.
\paragraph{Contrastive Loss for Positive and Negative Pairs}
The contrastive loss is bifurcated into two segments, addressing intra-class compactness and inter-class separability. The calculations include class weights to give different importance to samples from different classes:

\begin{itemize}
\label{eq:10}
\item Positive Pair Loss (\( L_{\text{pos}} \)): This component, calculated as:
\begin{equation}
\label{eq:11}
L_{\text{pos}} = -\sum_{i=1}^{N}\sum_{j=1}^{N} w_i \cdot \mathbb{1}_{\{y_i = y_j\}} \log \frac{\exp(\hat{x}_i^\top \hat{x}_j / \tau)}{\sum_{k=1}^{N} \exp(\hat{x}_i^\top \hat{x}_k / \tau)},
\end{equation}
focuses on reducing distances between samples from the same class, where \( w_i \) represents the class weight for sample \( i \).
\item Negative Pair Loss (\( L_{\text{neg}} \)): Computed as:
\begin{equation}
L_{\text{neg}} = -\sum_{i=1}^{N}\sum_{j=1}^{N} w_i \cdot \mathbb{1}_{\{y_i \neq y_j\}} \log \frac{\exp(-\hat{x}_i^\top \hat{x}_j / \tau)}{\sum_{k=1}^{N} \exp(-\hat{x}_i^\top \hat{x}_k / \tau)},
\end{equation}
it aims to increase distances between samples from different classes, also incorporating class weights \( w_i \) to balance the contribution from different classes.
\end{itemize}

\paragraph{Cluster Assignment Loss}
This component, represented by \( L_{\text{cluster}} = -\sum_{c=1}^{C} p_c \log p_c \), encourages a uniform distribution of samples across clusters, mitigating the effects of class imbalance by promoting balanced clustering.
\paragraph{Total Loss Combination}
The aggregate loss (\( L_{\text{CCL}} = L_{\text{pos}} + L_{\text{neg}} + L_{\text{cluster}} \)) synergizes these components, offering a comprehensive framework to address class imbalance by emphasizing both relative positioning of samples in the feature space and equitable cluster formation.

In the training process, we optimize both classification and clustering objectives by combining \textit{ContrastiveClusteringLoss} (denoted as $L_{\text{CCL}}$, which encompasses positive pair loss, negative pair loss, and cluster assignment loss) with CrossEntropyLoss (denoted as $L_{\text{CE}}$). The final loss is computed as:
\begin{equation}
L_{\text{combined}} = \alpha L_{\text{CE}} + \beta L_{\text{CCL}},
\end{equation}
where $\alpha$ and $\beta$ represent the relative contributions of each loss term. Through hyperparameter analysis, the ratio is set to 1:1 in the experiments. The experimental results are presented in the \ref{Hyperparameter analysis} section.

\paragraph{Theoretical Proof and Implications}
The theoretical proof of the effectiveness of this loss function involves demonstrating how it optimally adjusts the feature space to enhance both intra-class similarity and inter-class dissimilarity. Gradient analysis of the loss function with respect to the embeddings shows that the gradients encourage the model to reduce distances between similar samples (positive pairs) and increase distances between dissimilar samples (negative pairs), while also pushing towards balanced cluster sizes.

Empirical validation is essential to complement the theoretical analysis. This involves demonstrating that models trained with this loss function perform better on class-imbalanced datasets compared to models trained with conventional loss functions, particularly in terms of balanced class representation and discrimination.

In conclusion, the \textit{ContrastiveClusteringLoss} function emerges as a significant advancement in tackling the perennial issue of class imbalance in ML. By integrating feature representation learning with equitable cluster formation, and incorporating class weights, it addresses both the discriminative and representative aspects of learning in imbalanced scenarios. Algorithm \ref{training_alg} provides an overview of the key steps in our DL model for SSE.

\begin{algorithm}
\DontPrintSemicolon
\caption{Training Algorithm for Multi-Scale Dynamic Correlations SSE}
\label{training_alg}
\KwIn{Ship motion data $\mathcal{X} \in \mathbb{R}^{N \times F \times T}$, sea states $y \in \{0,...,6\}$, number of subsequences $s$, number of graph iterations $n$, learning rate $\eta$}
\KwOut{Sea state class prediction $\hat{y}$}
Preprocess $\mathcal{X}$ (normalize and segment data into time windows)\;
Initialize parameters of the Time Dimension Factorization (TDF), Dynamic Graph-like Learning (DGL), Combined Loss (CL), and Classification modules\;
\Repeat{convergence}{
    $\{x_1, ..., x_s\} \leftarrow \textit{TDF}(\mathcal{X})$\;
    $X^T \leftarrow \textit{MLP}(x_1, ..., x_s)$\;
    $X_{cnn} \leftarrow \textit{1D-CNN}(\mathcal{X})$\;
    $X_e \leftarrow \textit{Cat}(X^T, X_{cnn})$\;
    \For{iteration $i \in \{1, ..., n\}$}{
        $e_{i,j} \leftarrow {f_{n \to e}}(\textit{Cat}[X_{e,i}, X_{e,j}]), \forall (i,j)$\;
        $In_{i,j} \leftarrow \textit{sigmoid}(g(e_{i,j})), \forall (i,j)$\;
        $\mathbf{x}_i^{(2)} \leftarrow \sum_{j \neq i} e_{i,j} \cdot In_{i,j}, \forall i$\;
    }
    $\mathbf{p} \leftarrow \textit{ClassificationModule}(\mathbf{x}_i^{(2)})$\;
    $\mathcal{L}_{\text{combined}} \leftarrow \textit{ComputeCL}(\mathbf{x}_i^{(2)}, y)$\;
    Update parameters using gradient descent: $\Theta \leftarrow \Theta - \eta \nabla_{\Theta} \mathcal{L}_{\text{CL}}$\;
}
$\hat{y} \leftarrow \argmax(\mathbf{p})$\;
\end{algorithm}

\section{EXPERIMENTS}
\label{experiments}
The model employed PyTorch (v.2.4.1), a DL framework. Experiments were conducted on an NVIDIA GeForce RTX 4090 GPU-equipped server. The model optimization was performed using the Adam algorithm. Parameters included a learning rate of 5e-3, batch size of 128, and 500 learning epochs.
\subsection{Datasets and evaluation metrics}
\paragraph{Ship motion datasets}

The dataset for this study was obtained from the Marine System Simulator (MSS) toolbox \cite{perez2009matlab}, a well-known tool for simulating dynamic operations under various sea conditions. This research covers both global and North Atlantic sea regions. As detailed in Table \ref{table_seacode}, the cumulative probability of encountering a sea state between 0 and 6 exceeds 98\% in both areas, with less frequent sea states excluded from the analysis. The first three conditions were grouped into "Sea State 1" due to their similarities, while the remaining conditions were classified as Sea States 2 through 5. To accurately represent the probability of each condition, we adjusted the sample sizes accordingly. Two separate datasets were created to reflect geographic variations in occurrence probabilities: one for global data and another for the North Atlantic. For the simulation, inputs included heave velocity, pitch angle, pitch velocity, and yaw angle. The resulting ship motion data were split into a training set (70\%), a validation set (10\%), and a test set (20\%) to ensure proper distribution for model evaluation.

\paragraph{Benchmark Datasets} For performance comparison, we selected the publicly accessible UEA time series classification dataset \cite{bagnall2018uea}. This dataset includes data from various domains like human activity recognition, Electroencephalogram/electrocardiogram classification, and digit recognition. These datasets vary in dimensions and categories. We chose 14 datasets from the UEA collection, each exhibiting a broad range of cases, dimensions, and sequence lengths. The diversity and complexity of these datasets make them perfect for assessing our model's performance.

\paragraph{Evaluation metrics}
In this study, we use four metrics to evaluate the performance of the model. When analyzing the ship motion dataset, we take into account its inherent class imbalance. We use three key metrics: macro precision (P), macro recall (R), and macro F1 score (F1). These metrics can help us evaluate the model's performance in each class, regardless of the sample size. For the UEA dataset, where the number of samples in each category is consistent, we chose accuracy as the validation metric. The rationale behind introducing this dataset was to verify the generalization capability of our proposed model.

\begin{small} 
\begin{equation}
\begin{aligned}
\mathit{MacroPrecision} = \frac{1}{N}\sum\nolimits_{i = 1}^N {\frac{{TP}}{{TP + FP}}},
\end{aligned}
\end{equation}
\end{small} 

\begin{small} 
\begin{equation}
\begin{aligned}
\mathit{MacroRecall} = \frac{1}{N}\sum\nolimits_{i = 1}^N {\frac{{TP}}{{TP + FN}}},
\end{aligned}
\end{equation}
\end{small} 

\begin{equation}
\begin{aligned}
\mathit{MacroF1} = \frac{2 \times \mathit{Precision} \times \mathit{Recall}}{\mathit{Precision} + \mathit{Recall}},
\end{aligned}
\end{equation}
where $ TP $, $ FP $, $ FN $, and $ TN $ represent true positive, false positive, false negative, and true negative, respectively.

\begin{table}[htp]
\centering
\caption{Performance Metrics for Different Methods}
\label{tab:performance Metrics for Different Methods}
\begin{tabular}{l|ccc|ccc} \hline\hline
Methods & \multicolumn{3}{c}{World\_wide} & \multicolumn{3}{c}{North\_Atlantic} \\
\cmidrule(lr){2-4} \cmidrule(lr){5-7}
 & P & R & F1 & P & R & F1 \\ 
\midrule
MLP & 64.14 & 57,94 & 59.01 & 69.25 & 58.97 & 60.90 \\
CNN & 73.59 & 57.91 & 60.04 & 71.29 & 57.40 & 60.82 \\
TapNet & 72.48 & 62.14 & 64.31 & 73.76 & 63.25 & 65.13 \\
FCN & 75.67 & 74.39 & 74.85 & 74.06 & 77.25 & 75.38 \\
LSTM-FCNS & 78.22 & 72.30 & 74.46 & 74.03 & 75.27 & 74.61 \\ 
\hline\hline
CNN-SSE & 76.36 & 66.99 & 69.40 & 72.24 & 65.51 & 67.74 \\
SpectralSeaNet & 73.46 & 63.76 & 67.89 & 73.58 & 64.93 & 66.45 \\
Hybrid & 75.13 & 65.12 & 68.91 & 75.57 & 63.87 & 69.32 \\
DynamicSSE & 73.68 & 69.22 & 70.26 & 72.21 & 70.12 & 70.19 \\
SeaStateNet & 77.13 & 68.97 & 71.57 & 76.23 & 70.12 & 72.43 \\
SSENet & 78.46 & 71.79 & 74.02 & 78.47 & 72.39 & 74.18 \\
ImbalanceSSE & 78.68 & 78.22 & 78.38 & 77.01 & 74.46 & 75.46 \\
\textbf{TGC-SSE}  & \textbf{82.14} & \textbf{82.65} & \textbf{82.26} & \textbf{83.94} & \textbf{79.77} & \textbf{81.56} \\ \hline
\bottomrule
\end{tabular}
\end{table}

\begin{table*}[ht]
\caption{Accuracy Comparison of SOTA Methods Using 14 Benchmark Time Series Datasets}
\label{tab:benchmark}

\centering
\scalebox{1}{
\begin{adjustbox}{max width=\textwidth}
\begin{tabularx}{\textwidth}{l|l|*{10}{>{\centering\arraybackslash}X}{}}
\hline \hline
Dataset        &slice           & EDI    & DTWI        & DTWD        & MLSTM-FCNs  & WEASEL-MUSE & Negative Samples          & TapNet\cite{yoon2019tapnet}      & ShapeNet & TGC-SSE        \\ \hline
ArticularyWordRecognition  &8 & 0.970   & 0.980        & 0.987       & 0.973       & \textbf{0.990}        & 0.987       & 0.987       & 0.987    & 0.973       \\ \hline
AtrialFibrillation    &80    & 0.267  & 0.267       & 0.220        & 0.267       & 0.333       & 0.133       & 0.333       & 0.400      & \textbf{0.733}       \\ \hline
ERing            &5         & 0.133  & 0.133       & 0.133       & 0.133       & 0.133       & 0.133       & 0.133       & 0.133    & \textbf{0.167}       \\ \hline
FaceDetection     &8        & 0.519  & N/A         & 0.529       & 0.545       & 0.545       & 0.528       & 0.556       & 0.602    & \textbf{0.638}       \\ \hline
FingerMovements    &5       & 0.550   & 0.520        & 0.530        & 0.580        & 0.490        & 0.540        & 0.530        & 0.580     & \textbf{0.610}        \\ \hline
HandMovementDirection   &8  & 0.278  & 0.306       & 0.231       & 0.365       & 0.365       & 0.270        & 0.378       & 0.338    & \textbf{0.500}         \\ \hline
Heartbeat         &5        & 0.619  & 0.658       & 0.717       & 0.663       & 0.727       & 0.737       & 0.751       & \textbf{0.756}    & 0.722       \\ \hline
Libras         &5           & 0.833  & \textbf{0.894}      & 0.870        & 0.856       & 0.878       & 0.867       & 0.850        & 0.856    & 0.872      \\ \hline
NATOPS     &3               & 0.850   & 0.850        & 0.883       & 0.889       & 0.870        & \textbf{0.944}       & 0.939       & 0.883    & 0.906       \\ \hline
PEMS-SF         &8          & 0.705  & 0.734       & 0.711       & 0.699       & N/A         & 0.688       & 0.751       & 0.751    & \textbf{0.798}       \\ \hline
PenDigits       &1          & 0.973  & 0.939       & 0.977       & 0.978       & 0.948       & 0.983       & 0.980        & 0.977    & \textbf{0.987}        \\ \hline
SelfRegulationSCP1    &8    & 0.771   & 0.765       & 0.775       & \textbf{0.874}        & 0.710        & 0.846      & 0.652       & 0.782    & 0.802       \\ \hline
SelfRegulationSCP2    &8    & 0.483  & 0.533       & 0.539       & 0.472       & 0.460        & 0.556       & 0.550        & 0.578    & \textbf{0.594}       \\ \hline
StandWalkJump      &10       & 0.200    & 0.333       & 0.200         & 0.067       & 0.333       & 0.400         & 0.400         & \textbf{0.533}    & \textbf{0.533}       \\ \hline
\hline
Average accuracy     & N/A    & 0.582 & 0.609 & 0.593 & 0.597 & 0.599 & 0.615 & 0.628 & 0.654    & \textbf{0.703} \\ \hline
Wins/Ties      & N/A            & 0      & 1           & 0           & 1           & 1           & 1           & 0           & 2        & \textbf{9}          \\ \hline \hline
\end{tabularx}
\end{adjustbox}}
\end{table*}

\subsection{Baseline comparison of sea state estimation}
To fully verify the performance of the proposed model, we compare the state-of-the-art (SOTA) works in the domain of data-driven sea state estimation as well as some baseline models in the domain of time series classification. The details of the used methods are as follows:
\begin{enumerate}
\item \textbf{MLP} \cite{miao2018novel}: MLP finds extensive application in numerous ML tasks. In our study, we employed Miao’s approach and modified the model to include three layers, each with 100 hidden units.

\item \textbf{CNN} \cite{alfsen2020imu}: For the application of DL techniques in SSE, CNNs are frequently used as a benchmark. To enhance performance metrics for our dataset, we adjusted the model parameters and changed the number of filters in the CNN architecture.

\item \textbf{TapNet} \cite{yoon2019tapnet}: TapNet introduces a neural network enhanced with task-adaptive projection, leveraging a meta-learning approach to advance few-shot learning capabilities.

\item \textbf{FCN and LSTM-FCNs} \cite{karim2017lstm}: Fully Convolutional Networks (FCNs) utilize locally connected layers, including convolution, pooling, and upsampling. LSTM-FCNs, a well-established approach for time series classification, leverage a parallel architecture to simultaneously capture temporal and spatial features.

\item \textbf{CNN-SSE} \cite{kawai2021sea}: The design of CNN-SSE balances computation time and estimation accuracy. It includes two convolutional layers, two max-pooling layers, and four fully connected layers.

\item \textbf{SpectralSeaNet} \cite{cheng2020spectralseanet}: This DL model for SSE uses spectrograms as input, which are generated by applying the fast Fourier transform to ship motion data and then classified using a 2D CNN.

\item \textbf{Hybrid} \cite{han2021uncertainty}: A hybrid method for SSE incorporates uncertainty analysis by merging a data-driven model with a ship motion model. For this analysis, however, we focused solely on the data-driven model.

\item \textbf{DynamicSSE}: DynamicSSE introduces a GCN-based method to extract relationships between ship motion features.

\item \textbf{SeaStateNet} \cite{XU2019ICRA}: This is the first DL model for SSE, utilizing three branches to learn temporal, spatial, and frequency features from the motion data of dynamically positioned ships.

\item \textbf{SSENET} \cite{cheng2020novel}: This model introduces a densely connected convolutional neural network designed to enhance sea state estimation by leveraging ship motion data. This deep network simultaneously estimates both wave height and wave characteristics.

\item \textbf{ImbalanceSSE} \cite{cheng2023novel}: This model proposes a model integrating multi-scale and cross-scale feature learning modules with a prototype classifier to effectively address feature representation and data imbalance.
\end{enumerate}

The data in Table \ref{tab:performance Metrics for Different Methods} show the performance of different methods on two ship motion datasets, namely the World Wide dataset and the North Atlantic dataset. These three metrics reflect the effectiveness of the methods in classifying class-imbalanced ship motion data. It can be seen from the table that our proposed method (TGC-SSE) achieves the best performance on both datasets, indicating that our method can effectively extract the features of ship motion data and optimize the classification results using the dynamic graph-like structure and the contrastive clustering loss function. Compared with our method, other methods have a performance gap, especially the baseline models such as MLP, CNN and TapNet, and the SSE domain models such as CNN-SSE and SpectralSeaNet. Our analysis revealed that the baseline models do not take full advantage of the complexity and diversity of ship motion data. The SSE domain models lack specialised components for capturing the temporal dynamics and feature interactions that are unique to ship motion data, resulting in a suboptimal characterisation of the intrinsic properties of the data. With the exception of DynamicSSE, no model discusses the correlation between ship motion data features, and with the exception of ImbalanceSSE, no model focuses on the class imbalance features inherent to ship motion data. Among them, the method we proposed improved the F1 score of the World\_wide dataset by 4.96\% compared to ImbalanceSSE, which performed the best in the comparative experiment, and improved the F1 score of the North\_Atlantic data by 8.08\%. In summary, our model demonstrates superior performance and robustness in dealing with class-imbalanced ship motion data. This is attributed to our time dimension factorisation module, dynamic graph-like learning module, and contrastive clustering loss function. Our research provides a new and effective DL method for SSE.

\subsection{Comparisons with the SOTA methods on UEA datasets}
In order to demonstrate the generalization ability of our model as well as to evaluate our model more rigorously, we have selected 14 datasets from the UEA Multivariate Time Series Classification Archive, which contain a wide range of problems, including variable-length sequences. First of all we would like to characterize the size selection of slice and Sampling, first of all this number is selected as a factor of the last dimension of the data. In our model to deal with the ship movement data we used 8, so when dealing with this data, if 8 is a factor of the third dimension of the data set then slice = 8, if it is not a factor of his number is chosen similar to 8, for example, 3, 5, 10. But this is not necessarily the highest accuracy of our model to handle these datasets, the accuracy will go up further as we adjust the size of slice, learning rate and other parameters. We will verify our claims with one of these datasets later.

We evaluate the performance of our proposed model against eight baseline methods as follows:

\begin{enumerate}
\item \textbf{EDI, DTWI, and DTWD}: Enhanced Deep Interaction enhances interaction between different layers to improve feature extraction and classification accuracy; Dynamic Time Warping with Interpolation extends DTW by incorporating interpolation to handle time series data with different lengths; and Dynamic Time Warping Distance measures the similarity between two time series by calculating the optimal match, though it does not satisfy the triangle inequality \cite{bagnall2018uea, shokoohi2015non}.

\item \textbf{MLSTM-FCN}: This innovative DL model integrates LSTM and FCN to effectively capture both temporal dependencies and local features, thus improving classification performance \cite{karim2019multivariate}.

\item \textbf{WEASEL-MUSE}: This well-known model combines sliding window techniques and feature selection methods, using a bag-of-words model and multivariate symbolization to extract and filter features for classification. \cite{schafer2017multivariate}.

\item \textbf{Negative Samples}: This paper proposes an unsupervised scalable representation learning method, utilizing Causal Dilated Convolutions and a Triplet Loss Function to learn general embeddings for multivariate time series \cite{franceschi2019unsupervised}.

\item \textbf{ShapeNet}: This shapelet-based time series classification model embeds shapelets into a unified space and combines neural networks for classification \cite{li2021shapenet}.
\end{enumerate}

\begin{figure} [htp]
        \centering
		\includegraphics[width=0.42\textwidth]{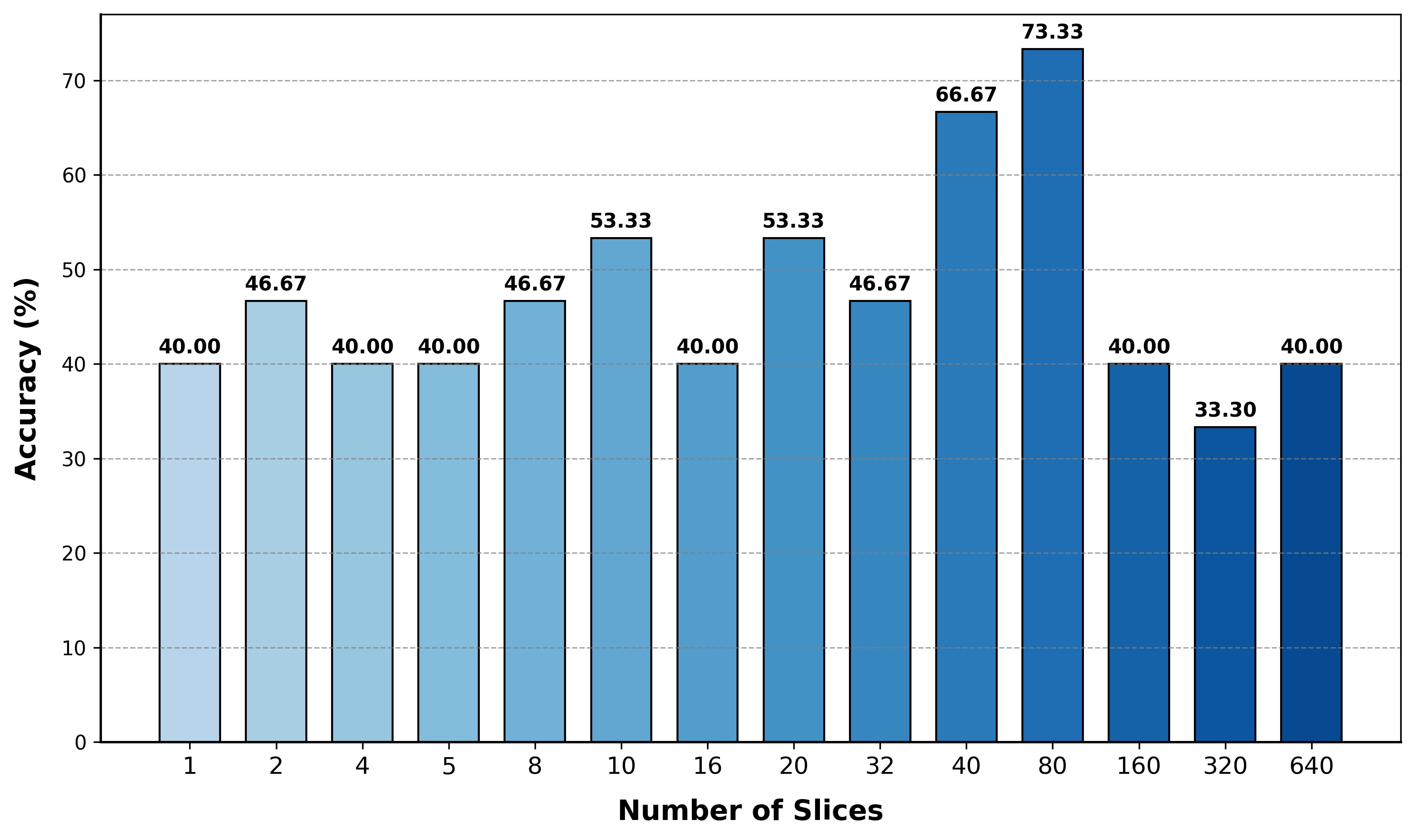}
		\vspace{-5pt}
        	\caption{Sensitivity Analysis on AtrialFibrillation Dataset.}
		\label{fig:sa}
		\vspace{-5pt}
\end{figure}

As shown in Table \ref{tab:benchmark}, in 14 public datasets, our model shows higher accuracy on the classification task compared to the other baseline models we listed. In the 14 datasets, our model classifies 9 datasets with the highest accuracy and in terms of average accuracy, our model improves 7.48\% compared to ShapeNet, which is ranked second, and 20.79\% compared to EDI, which has the lowest average accuracy. Our model performs very well on the dataset AtrialFibrillation, outperforming the second ranked method by 33.3\%. We did a sensitivity analysis experiment on this dataset. According to \cite{bagnall2018uea}, the dimensions of this dataset are (15, 2, 640). We performed slicing operations along the third dimension with slice and sampling factors chosen as divisors of 640. The slice factors and corresponding experimental results are listed in Fig. \ref{fig:sa}.

As can be seen from the bar chart, of the 14 factors in 640, the slice number of 80 has the highest accuracy, which is an improvement of 120.21\% compared to the lowest accuracy observed when the slice number is 320. This observation significantly highlights the data redundancy issue discussed in the first challenge. Sensitivity analysis further demonstrates that an appropriate number of slices and sampling rate positively influence the model's ability to learn features. This also explains why the AtrialFibrillation dataset chose a slice count and sampling rate of 80, which differs markedly from those selected for other datasets. In future research, we are interested in exploring how to dynamically set these parameters for different datasets or whether using different values for these parameters might yield better results.

\subsection{Comparison with class imbalance loss functions}

To further evaluate the proposed model, we compare it with five loss functions designed to handle imbalanced classification. Below is a detailed description of these loss functions:

\begin{enumerate}
\item \textbf{BalanceSoftmax}: This extension of Softmax normalizes the sample count in each category, making the weight of each category inversely proportional to its frequency in the overall dataset \cite{ren2020balanced}.

\item \textbf{Focalloss}: It enhances the model's attention on underrepresented categories by assigning higher weights to hard-to-classify samples while reducing the weights for easily classified ones \cite{lin2017focal}.

\item \textbf{Class-balanced}: This variation of focal loss introduces the concept of effective sample size. It considers both the sample count and classification difficulty to adjust loss weights, ensuring equivalent contribution from each class \cite{cui2019class}.

\item \textbf{WeightedCE}: This modification of Cross-Entropy Loss balances category losses by assigning fixed weights to each category, typically inversely proportional to their frequencies in the dataset \cite{mao2023cross}.

\item \textbf{LDAMLoss} (label-distribution-aware margin): This loss function aims to improve generalization on less frequent classes by encouraging larger margins for minority classes, thereby reducing overfitting and enhancing model performance on imbalanced datasets. We follow the same settings as in \cite{cao2019learning}.
\end{enumerate}

\begin{table}[ht]
\centering
\setlength{\tabcolsep}{1.6pt} 
\caption{Comparison of Proposed and Imbalanced Loss Functions for Ship Motion Data}
\begin{tabular}{@{}l|>{\centering\arraybackslash}p{1cm}>{\centering\arraybackslash}p{1cm}>{\centering\arraybackslash}p{1cm}|>{\centering\arraybackslash}p{1cm}>{\centering\arraybackslash}p{1cm}>{\centering\arraybackslash}p{1cm}@{}}
\hline \hline
Loss & \multicolumn{3}{c}{World\_wide} & \multicolumn{3}{c}{North\_Atlantic} \\
\cmidrule(lr){2-4} \cmidrule(lr){5-7}
 & P & R & F1 & P & R & F1 \\ 
\midrule
BalancedSoftmax & 63.78 & 65.58 & 64.50 & 63.15 & 63.03 & 62.87 \\
FocalLoss & 75.83 & 75.41 & 75.45 & 73.84 & 75.97 & 74.79 \\
Class-Balanced & 79.69 & 76.21 & 77.66 & 75.83 & 77.10 & 76.42 \\
WeightedCE & 77.35 & 77.67 & 77.40 & 76.38 & 79.47 & 77.61 \\
LDAMLoss & 80.53 & 77.61 & 78.74 & 79.07 & 77.68 & 78.22  \\
\textbf{CombinedLoss}  & \textbf{82.14} & \textbf{82.65} & \textbf{82.26} & \textbf{83.94} & \textbf{79.77} & \textbf{81.56} \\ \hline
\bottomrule
\end{tabular}
\label{tab:performance}
\end{table}

Table \ref{tab:performance} shows that the proposed Combined loss function outperforms other loss functions in training learning for ship motion data. By observing and comparing the F1 scores, on the World\_wide dataset, it improves by about 4.47\% compared to the best metric, LDAMLoss, and about 27.55\% compared to the lowest metric, BalancedSoftmax. On the North\_Atlantic dataset, it improves by about 4.27\% compared to the best metric, LDAMLoss, and by about 29.63\% compared to the lowest metric, BalancedSoftmax. We analyze that the lower performance of BalancedSoftmax may be due to overfitting caused by the large number of parameters. This also highlights the effectiveness of the Combined loss function in addressing the problem of class imbalance.

\subsection{Ablation study}

In order to evaluate the efficacy of each component of our proposed method, we designed four variants for comparison. The ``TDF+DGL+CL'' variant is our proposed complete model and uses our proposed Combined loss function. The ``TDF+DGL+CE'' variant simply removes our proposed contrastive clustering loss function and trains with the cross-entropy loss function. The ``TDF+DGL(no concat)+CL'' variant means that the features after the TDF module are no longer concatenated with the features after the DGL module. The ``TDF+CL'' variant means that we have removed the DGL module, and the ``DGL+CL'' variant means that we have removed the TDF module.

\begin{figure}[htp]
     \centering
     \begin{subfigure}[b]{0.24\textwidth}
         \centering
             \includegraphics[width=\textwidth]{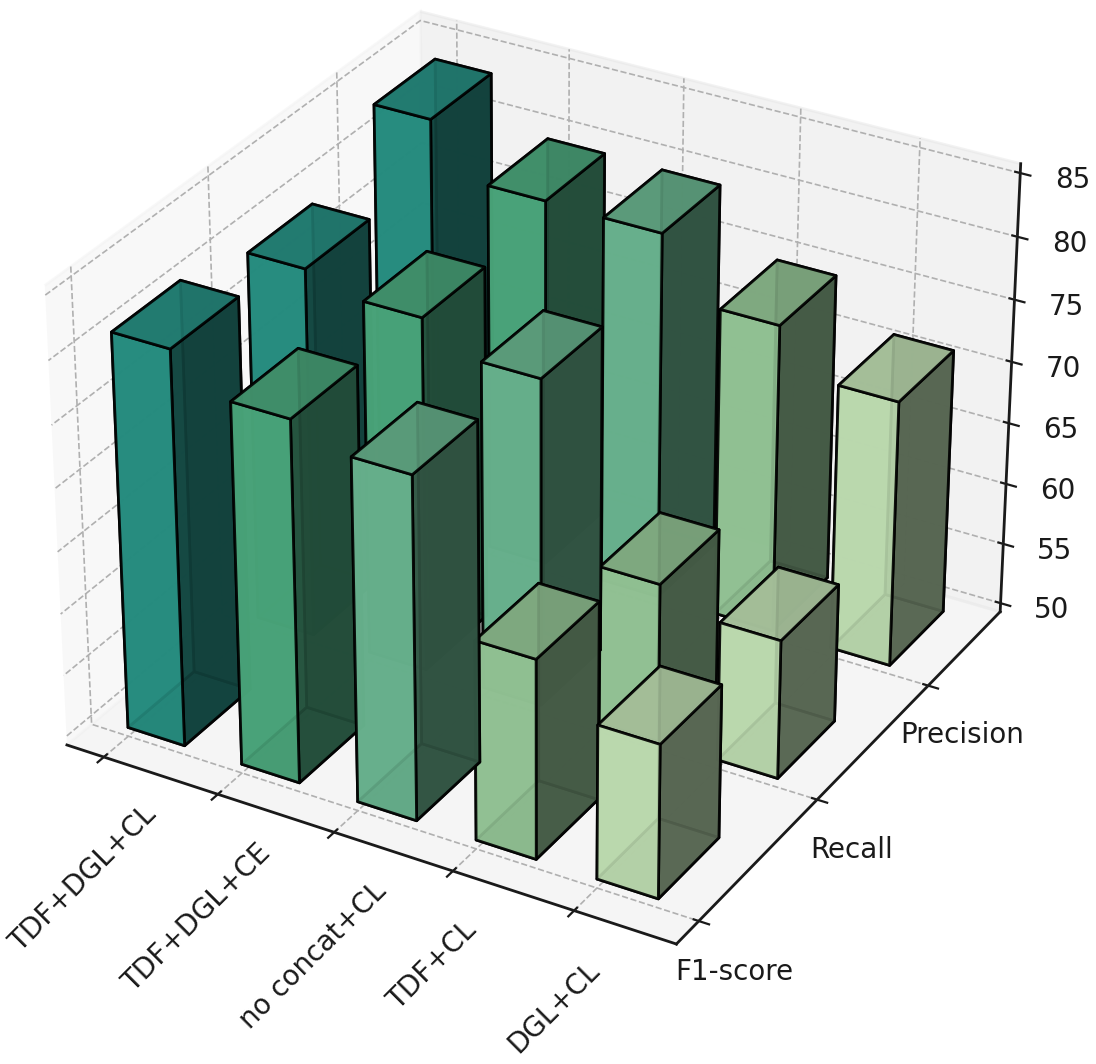}
         \caption{World wide}
         \label{fig:world wide xiaorong}
     \end{subfigure}
     \hfill
     \begin{subfigure}[b]{0.24\textwidth}
         \centering
         \includegraphics[width=\textwidth]{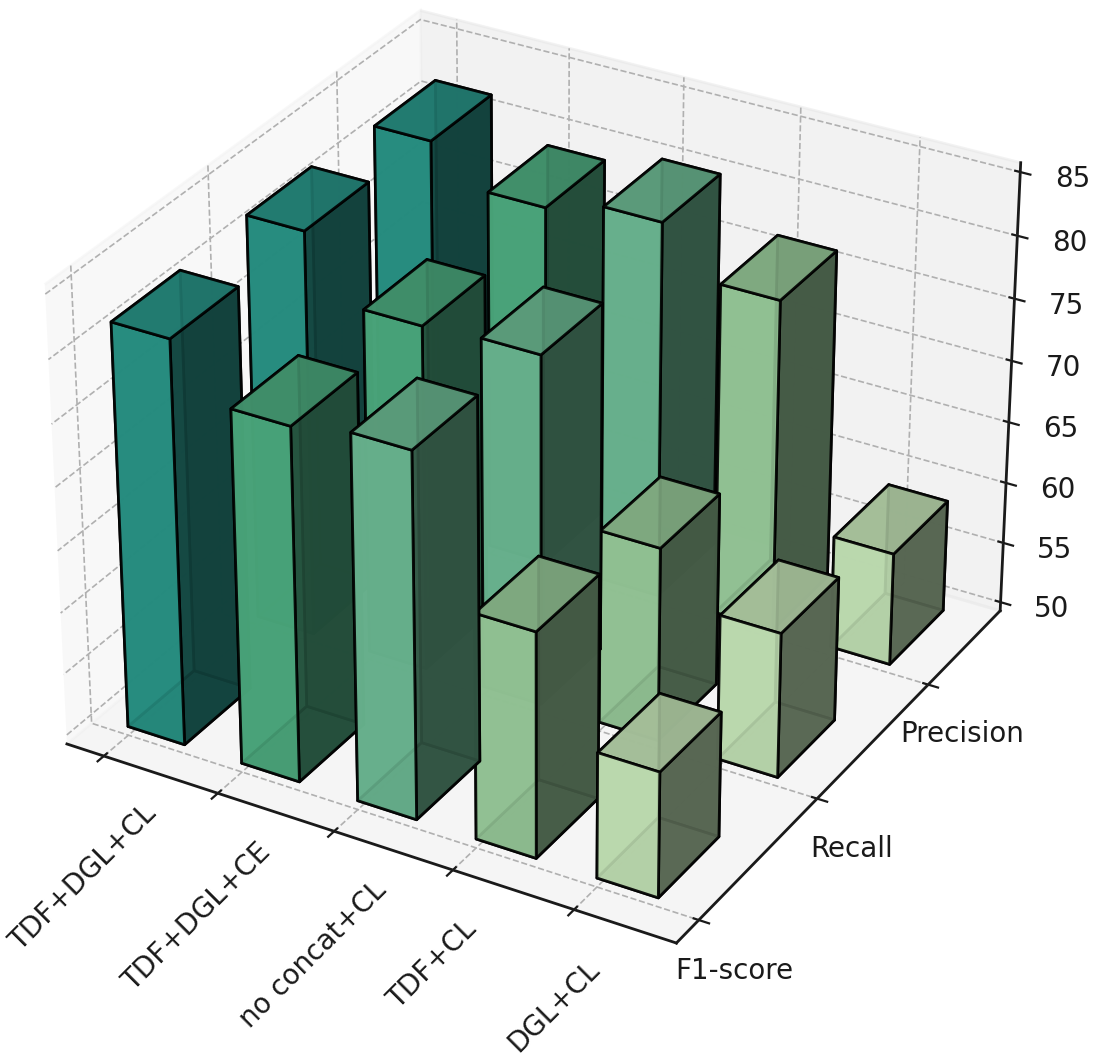}
         \caption{North Atlantic}
         \label{fig:north atlantic xiaorong}
     \end{subfigure}
   \caption{Ablation study.}
        \label{fig:imbalance_seastate}
\end{figure}

Figure \ref{fig:world wide xiaorong} and Figure \ref{fig:north atlantic xiaorong} show the ablation experiments in 3D bar charts. It is worth noting that if we delete two key modules of the proposed model, the performance decreases significantly. The analysis takes the World\_Wide dataset as an example. When the TDF module is removed, the F1 score drops by 27.04\%, and when the DGL module is removed, the F1 score drops by 17.35\%. This is enough to prove that our method extracts useful features from both the time and channel dimensions. Connecting the two modules and performing a concatenation operation results in a significant improvement in the metric, and not performing a concatenation operation results in a smaller decrease in performance. However, when only the cross-entropy loss is retained in the loss function, the F1 score decreases by 4.91\%, proving that the contrastive clustering loss can effectively deal with class imbalance. In short, these experimental results clearly demonstrate the importance of each component in our model and its impact on the final performance.

\subsection{Sensitivity analysis of temporal factorization}

\begin{figure}[htp]
     \centering
     \begin{subfigure}[b]{0.24\textwidth}
         \centering
             \includegraphics[width=\textwidth]{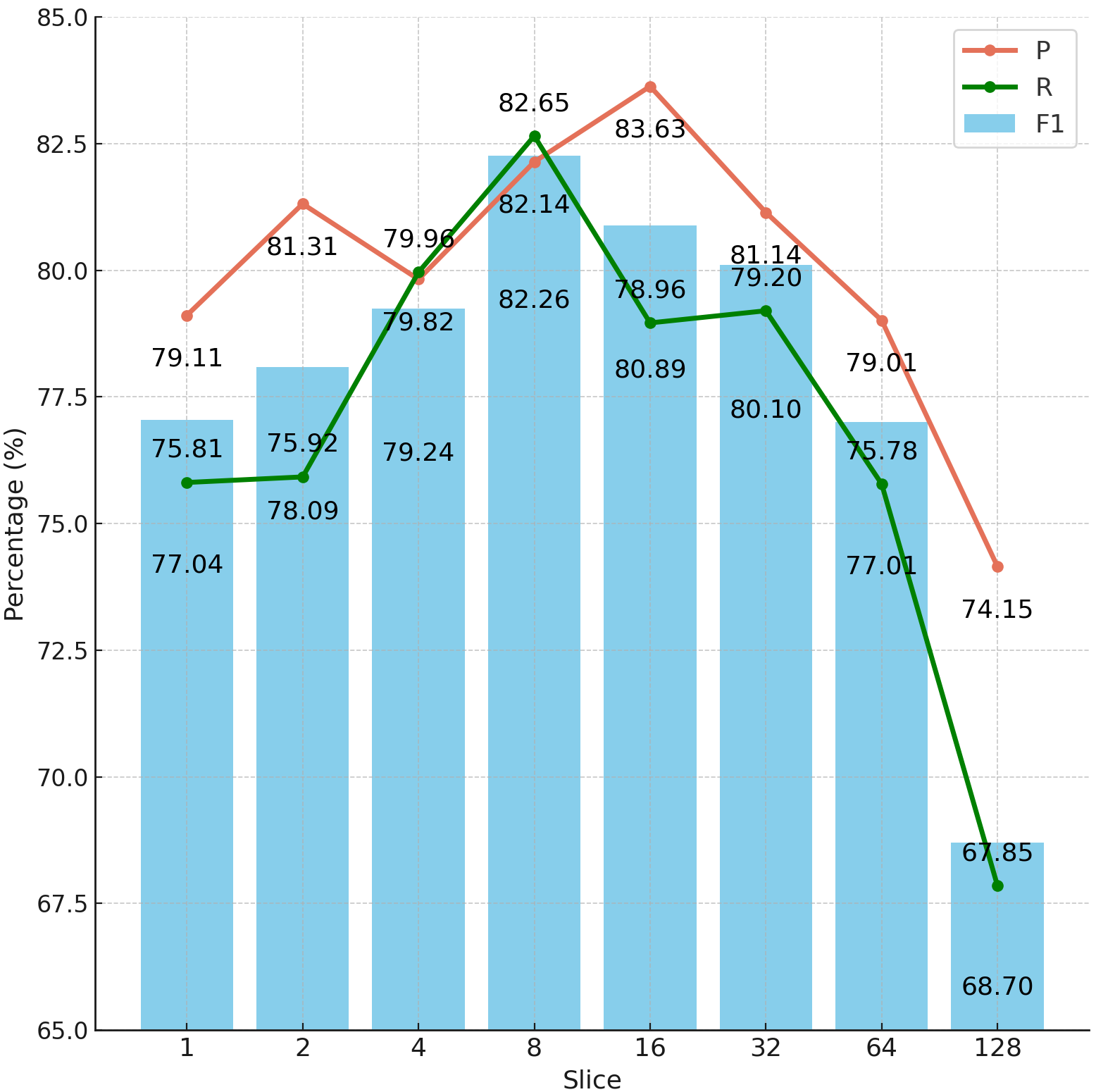}
         \caption{World wide}
         \label{fig:world wide sa}
     \end{subfigure}
     \hfill
     \begin{subfigure}[b]{0.24\textwidth}
         \centering
         \includegraphics[width=\textwidth]{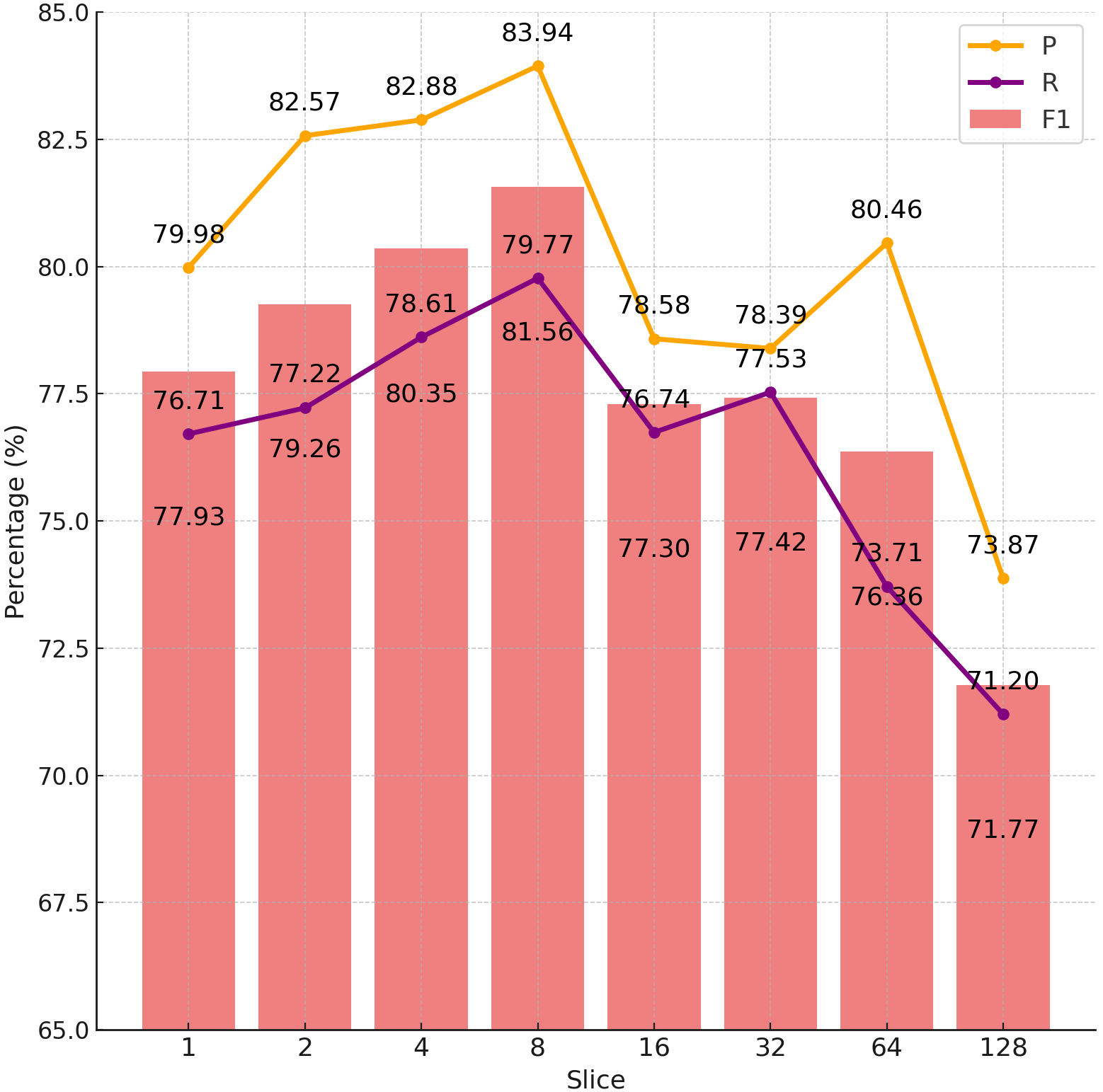}
         \caption{North Atlantic}
         \label{fig:north atlantic sa}
     \end{subfigure}
   \caption{Sensitivity analysis}
        \label{fig:Sensitivity analysis}
\end{figure}

To assess the impact of different slice numbers on model performance, we conducted a sensitivity analysis by varying the number of slices from 1 to 128 (1, 2, 4, 8, 16, 32, 64, 128). As shown in Fig \ref{fig:Sensitivity analysis}, smaller slice numbers (e.g., 1, 2) enable the model to capture broader temporal relationships, making it more effective at handling patterns with longer time spans. However, this may introduce redundant information when the time series exhibits many fine-grained variations. Conversely, larger slice numbers (e.g., 64, 128) allow the model to capture more detailed temporal dependencies by processing smaller segments of the sequence. While this is beneficial for detecting high-frequency temporal changes, it may lead to overfitting on subtle, irrelevant patterns. Our sensitivity analysis indicates that a slice number of 8 provides the best balance between capturing coarse and fine-grained temporal information, leading to improved generalization in multiple sea state classification tasks.

\subsection{Hyperparameter analysis}
\label{Hyperparameter analysis}

\begin{figure}[H]
    \centering
    \begin{subfigure}[b]{0.24\textwidth}
        \centering
        \includegraphics[width=\textwidth]{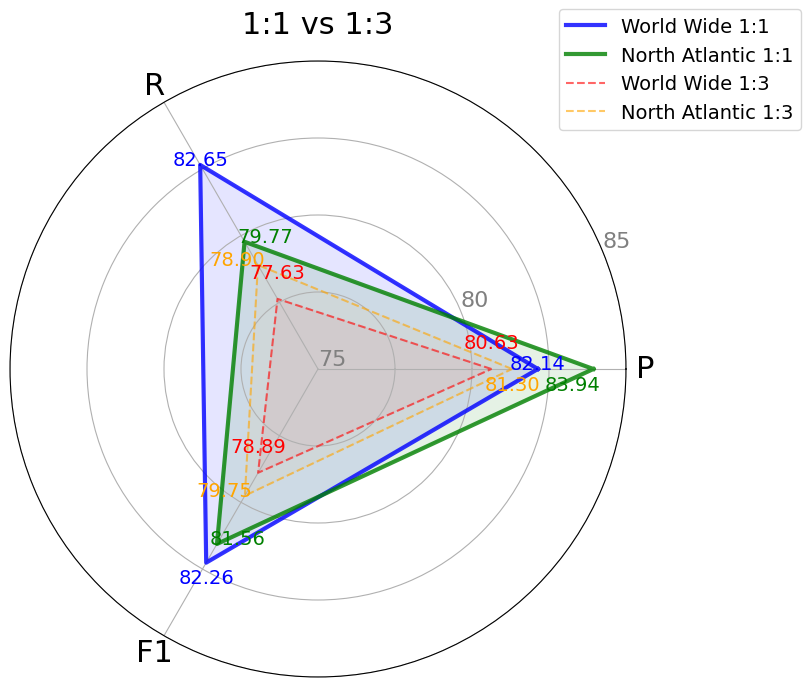}
        \label{fig:1：3}
    \end{subfigure}
    \hfill
    \begin{subfigure}[b]{0.24\textwidth}
        \centering
        \includegraphics[width=\textwidth]{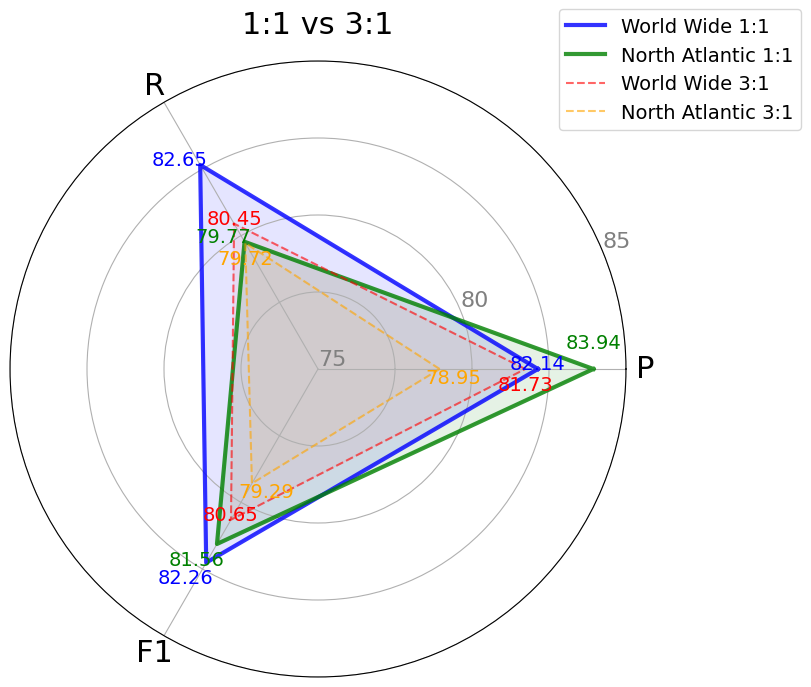}
        \label{fig:3：1}
    \end{subfigure}
    \caption{$\alpha L_{\text{CE}} : \beta L_{\text{CCL}}$.}
    \label{fig:CL}
\end{figure}

In this section, we analyze the impact of the hyperparameters \(\alpha\) and \(\beta\) on model performance, focusing on precision (P), recall (R), and F1 score. We evaluated different ratios of \(\alpha : \beta\), specifically 1:1, 1:2, 2:1, 3:1, and 1:3, across two sea state datasets. The results are visualized in the radar plot shown in Figure \ref{fig:CL}.

\begin{figure}[H]
    \centering
    \begin{subfigure}[b]{0.24\textwidth}
        \centering
        \includegraphics[width=\textwidth]{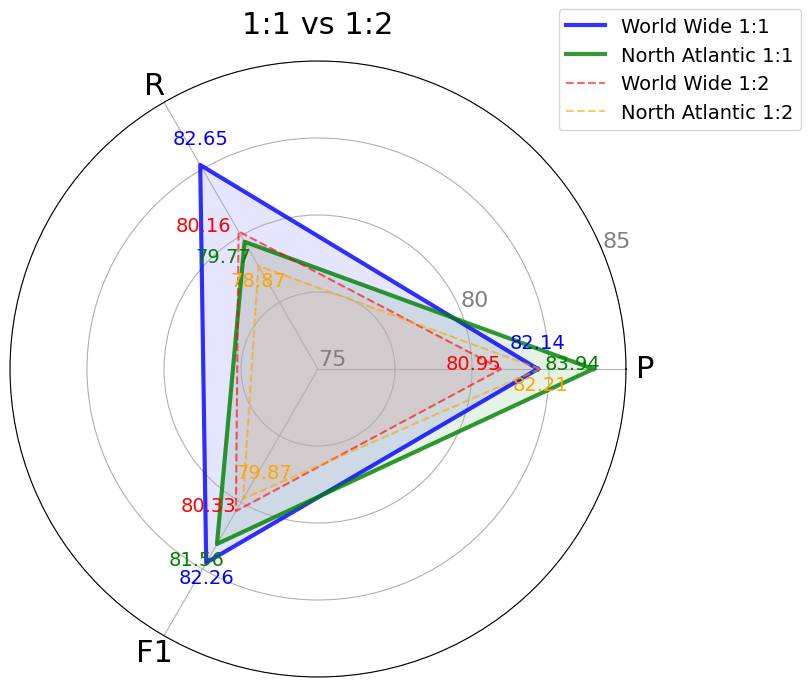}
        \label{fig:1:2}
    \end{subfigure}
    \hfill
    \begin{subfigure}[b]{0.24\textwidth}
        \centering
        \includegraphics[width=\textwidth]{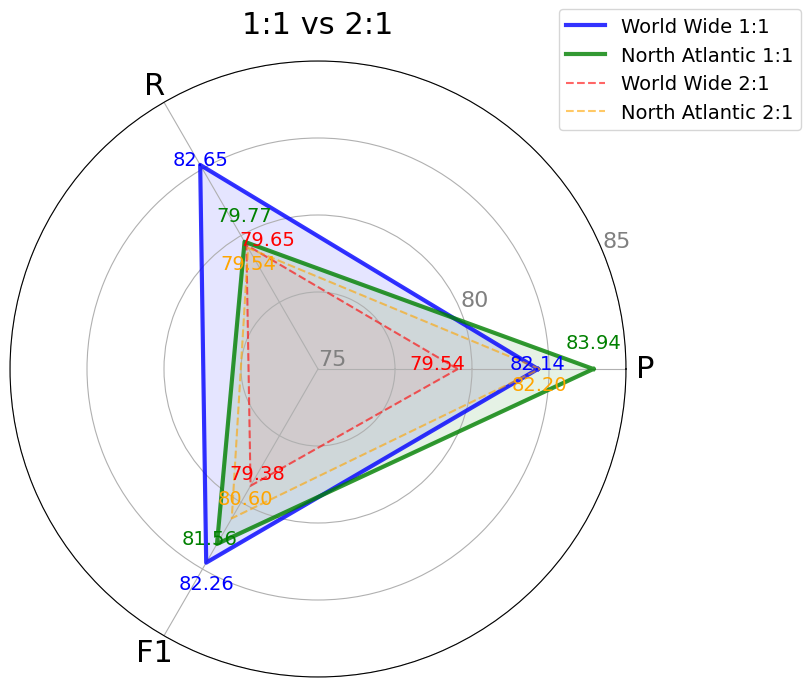}
        \label{fig:2：1}
    \end{subfigure}
\end{figure}

\vspace{-10pt} 

The experimental results indicate that the 1:1 ratio of \(\alpha : \beta\) consistently achieves the best performance across both datasets. For the World\_wide dataset, the 1:1 ratio achieved a recall of 82.65\%, which is an improvement of 6.47\% over the 1:3 ratio. Similarly, in the North\_Atlantic dataset, the 1:1 ratio achieved a precision of 83.94\%, which is 6.32\% higher than the 3:1 ratio.

The use of cross-entropy loss combined with contrastive clustering loss plays a crucial role in achieving balanced performance across different classes. The 1:1 ratio ensures that both \(\alpha\) and \(\beta\) contribute equally, allowing the model to learn representative features without bias towards any specific class. This balanced weighting configuration improves the model's generalization capability and stability, resulting in higher F1 scores across different datasets.

In contrast, other ratios, such as 1:2 or 3:1, lead to an imbalance in the contribution of \(\alpha\) and \(\beta\), which negatively impacts model performance. For example, in the World\_wide dataset, the 3:1 ratio results in a recall reduction to 80.45, indicating that overemphasizing one hyperparameter can lead to inadequate coverage of certain features, thereby reducing recall and ultimately affecting the F1 score.

The radar plot in Figure \ref{fig:CL} highlights that the 1:1 ratio provides the most balanced performance across all metrics, which is crucial for applications requiring consistent precision and recall. Balanced weighting schemes generally enhance generalization and stability. Therefore, \(\alpha : \beta = 1:1\) is the most suitable configuration for our model, ensuring robustness across diverse datasets.

\subsection{Classification performance visualization analysis}

\begin{figure}[H]
    \centering
    \begin{subfigure}[b]{0.24\textwidth}
        \centering
        \includegraphics[width=\textwidth]{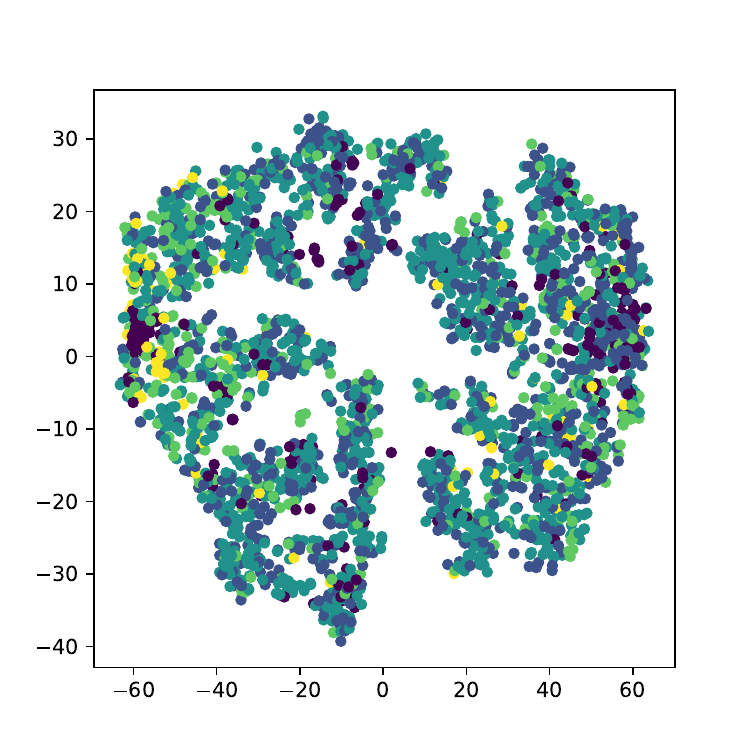}
        \caption{World wide}
        \label{fig:world wide zhexian1}
    \end{subfigure}
    \hfill
    \begin{subfigure}[b]{0.24\textwidth}
        \centering
        \includegraphics[width=\textwidth]{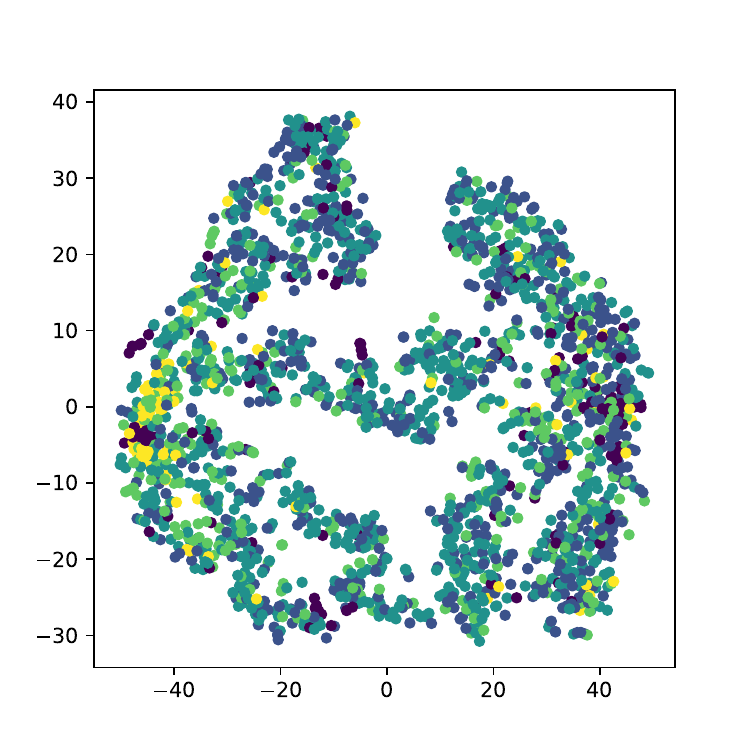}
        \caption{North Atlantic}
        \label{fig:north atlantic zhexian1}
    \end{subfigure}
    \caption{Original signal.}
    \label{fig:original signal}
\end{figure}

\begin{figure}[H]
    \centering
    \begin{subfigure}[b]{0.24\textwidth}
        \centering
        \includegraphics[width=\textwidth]{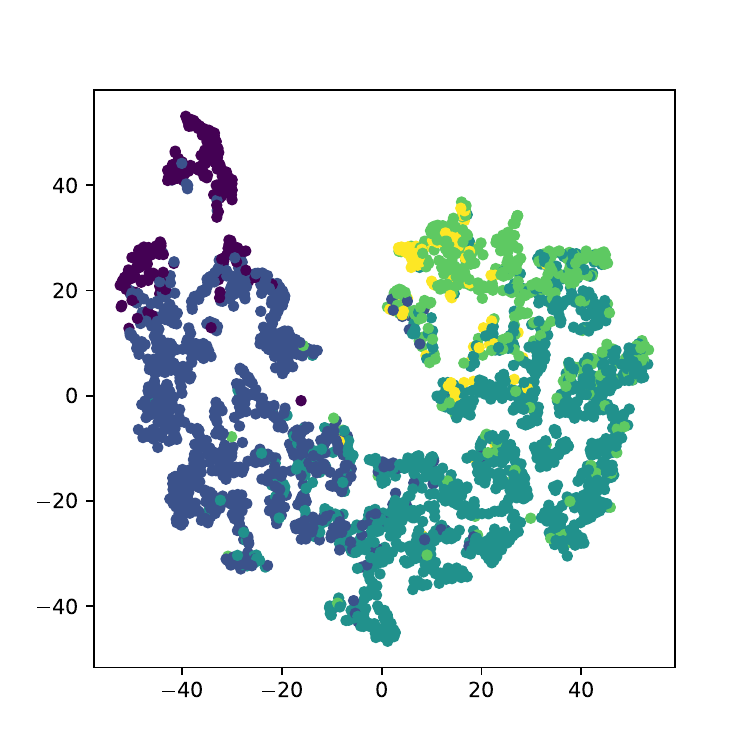}
        \caption{World wide}
        \label{fig:world wide zhexian2}
    \end{subfigure}
    \hfill
    \begin{subfigure}[b]{0.24\textwidth}
        \centering
        \includegraphics[width=\textwidth]{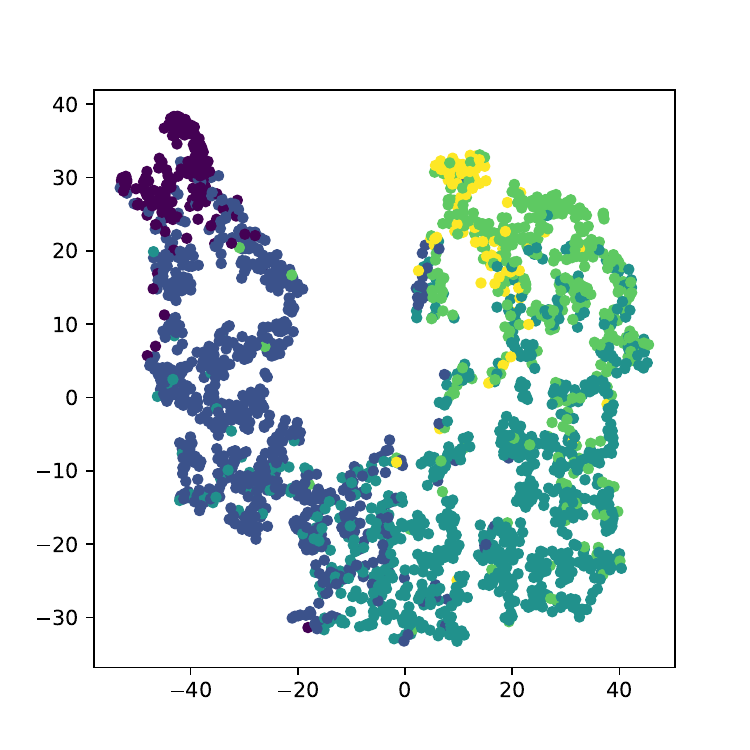}
        \caption{North Atlantic}
        \label{fig:north atlantic zhexian2}
    \end{subfigure}
    \caption{Cross Entropy Loss.}
    \label{fig:Cross Entropy Loss}
\end{figure}

In this section, we visualize and illustrate the original signal and the features processed by the two modules to emphasize the efficacy of our proposed model in a more intuitive way. We demonstrate the feature embedding visualization generated by the t-SNE algorithm using 2D images. Fig. \ref{fig:original signal} depicts the raw signal computation results for the two ship motion datasets, with solid circles indicating samples from the test set. Fig. \ref{fig:Cross Entropy Loss} illustrates the classification results after model training and cross-entropy loss function optimization.

\begin{figure}[H]
    \centering
    \begin{subfigure}[b]{0.24\textwidth}
        \centering
        \includegraphics[width=\textwidth]{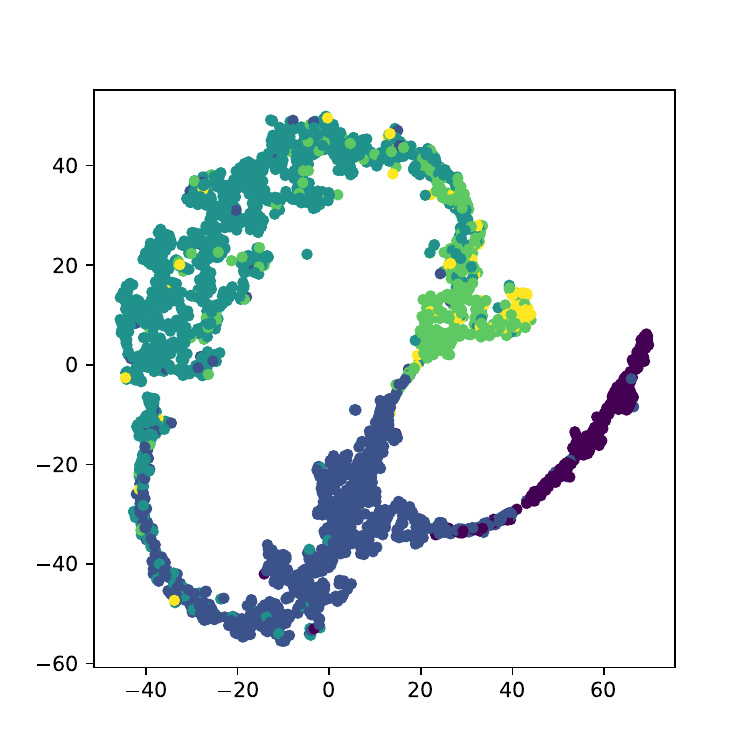}
        \caption{World wide}
        \label{fig:world wide zhexian3}
    \end{subfigure}
    \hfill
    \begin{subfigure}[b]{0.24\textwidth}
        \centering
        \includegraphics[width=\textwidth]{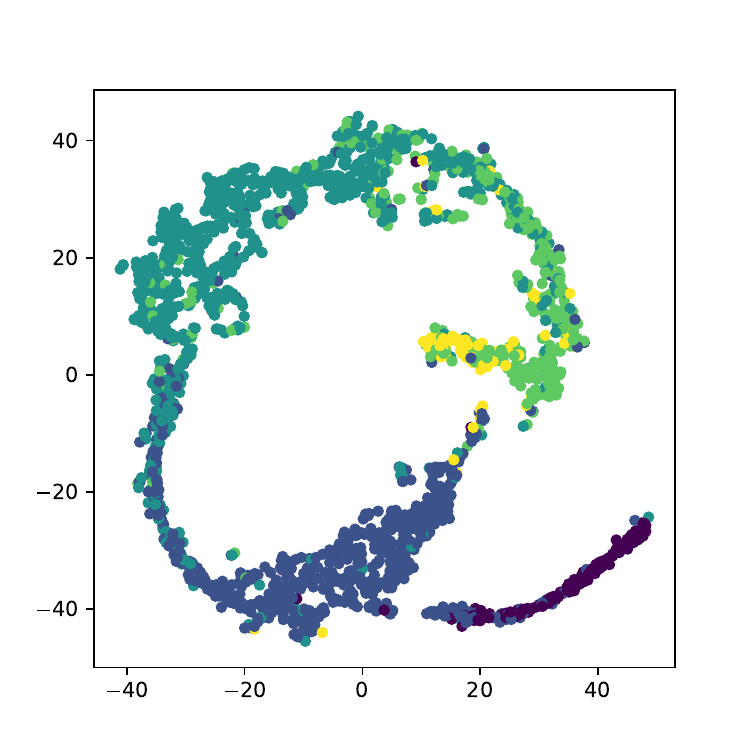}
        \caption{North Atlantic}
        \label{fig:north atlantic zhexian3}
    \end{subfigure}
    \caption{Contrastive Clustering Loss.}
    \label{fig:ContrastiveClusteringLoss}
\end{figure}

Given the category imbalance in the samples, we integrated the custom Contrastive clustering loss function with the cross-entropy loss function during model optimization. Comparing the visualization results of Fig. \ref{fig:Cross Entropy Loss} and Fig. \ref{fig:ContrastiveClusteringLoss}, it can be seen that the intra-class distance in Fig. \ref{fig:ContrastiveClusteringLoss} is reduced compared to Fig. \ref{fig:Cross Entropy Loss}, while the inter-class distance is increased at the same time. This indicates that by incorporating our designed loss function, the model is able to learn a better representation and improve the generalization ability, i.e., more accurate prediction on unseen data. And Figure \ref{fig:ContrastiveClusteringLoss} can also clearly show that our model design is an excellent model for SSE compared to Figure \ref{fig:original signal}.

\subsection{Discussion}

In this section, we delve into the implications of the experimental results of our proposed TGC-SSE model, highlighting its strengths and acknowledging its limitations. The experimental results demonstrate that our model exhibits superior classification performance compared to other DL models applied to sea state classification, proving the efficacy of our innovative approach in addressing the challenge.

\vspace{-20pt} 
\vspace{-20pt} 

We attribute this result to the inadequacy of existing models in adequately accounting for the class imbalance and multi-scale features inherent in ship motion data, as well as the neglect of temporal information and channel correlation, which are effectively addressed by the TGC-SSE model through the integration of a time-dimensional factorisation module, a dynamic graph-like learning module, and a contrastive clustering loss, resulting in enhanced generalisation and robustness.

Moreover, the model's excellent performance on 14 UEA multivariate time series datasets enhances its generality and generalisation. It is proficient in constructing dynamic graph structures and employing contrast clustering for balanced classification even when dealing with time series data with varying dimensionality, domain and class distributions. This study introduces a novel and effective DL approach for SSE.

However, our model also has some limitations and areas for improvement. First, the computational cost may be higher than some benchmark methods due to the need to construct dynamic graph structures for feature extraction. To mitigate this problem, we employ a 1D CNN to reduce the output dimension of the feature extraction module, thereby reducing the number of features processed by the dynamic graph construction module. In addition, we iteratively optimise the graph representation to minimise computational and memory overheads.

Second, model performance may be sensitive to certain hyperparameters such as slice size, learning rate, and number of iterations. These parameters need to be tuned to achieve optimal performance on different datasets. To simplify this process, we can later opt for adaptive methods such as automatic hyperparameter tuning or meta-learning, which allow the model to autonomously search for the best hyperparameter combination.

Finally, our model is currently limited to classification tasks and does not support regression tasks. To extend the applicability of the model, it may be necessary to train the model using regression or prediction loss functions (e.g., mean square error or root mean square error) for a wider range of SSE tasks.

In conclusion, our TGC-SSE model represents an innovative DL approach to the challenge of category imbalance in maritime condition classification. It efficiently extracts features from the temporal dimension, exploits the dynamic graph structure for fine-grained feature extraction, and employs contrastive clustering loss for balanced classification. The performance of the model on two ship motion datasets and 14 UEA datasets demonstrates its effectiveness and generality. Our study provides a new solution in the field of SSE and provides insights and directions for future research.

\section{Conclusion}
\label{conclusion}

The accurate and timely classification of sea states is pivotal for making informed navigational decisions under varying maritime conditions. Our research contributes to this critical area by introducing an innovative DL model capable of learning from imbalanced ship motion data and classifying sea states effectively. This capability not only enhances safety but also optimizes energy efficiency, aligning with the urgent demand for more sustainable and efficient solutions in the maritime industry. In this study, we have comprehensively evaluated our proposed TGC-SSE model, and the experimental results have demonstrated its superiority over benchmark methods. This innovation not only enriches the body of knowledge in our field but also holds promise for practical applications, offering new possibilities for autonomous ships and marking a significant step towards more responsive, sustainable, and safe navigation. While our model has shown promising results, there is scope for future work. For instance, extending the model's functionality to include regression tasks, thereby expanding its applicability in marine condition estimation. At the same time, the frequency domain characteristics of ship motion data are being explored, and the feasibility of extracting both time and frequency domain characteristics is being studied, in order to estimate sea conditions more promptly and accurately. In addition, adaptive methods for hyperparameter tuning are being further studied to enhance the model's performance in various datasets. In conclusion, our work lays an essential foundation for advancing towards a more sustainable and efficient maritime industry. The TGC-SSE model exemplifies the potential of DL in tackling the complex challenge of SSE, setting the stage for future innovations in maritime safety and navigation.


%



\ifCLASSOPTIONcaptionsoff
  \newpage
\fi

\bibliographystyle{IEEEtran}
\bibliography{Reference}

\typeout{get arXiv to do 4 passes: Label(s) may have changed. Rerun}

\end{document}